\documentclass[10pt,journal,compsoc]{IEEEtran}
\ifCLASSOPTIONcompsoc
  \usepackage[nocompress]{cite}
\else
  \usepackage{cite}
\fi
\usepackage{graphicx}
\ifCLASSINFOpdf
\else
\fi
\usepackage{amsmath}
\usepackage{amssymb}

\usepackage{booktabs}
\usepackage{makecell}
\usepackage{multirow}
\usepackage{array}
\usepackage{adjustbox}
\usepackage{tikz}

\usepackage{subfig}
\usepackage{threeparttable}
\usepackage{flushend}
\usepackage{bm}
\usepackage{stmaryrd}
\usepackage{xcolor}
\definecolor{xmudared}{RGB}{255, 102, 102}
\definecolor{xmudablue}{RGB}{102, 178, 255}
\definecolor{xmudabluedark}{RGB}{0, 0, 126}
\usepackage{hyperref}

\newcommand{\xs}[1]{\bm{x}_s^\text{#1}}
\newcommand{\ys}{\bm{y}_s^\text{3D}}
\newcommand{\xt}[1]{\bm{x}_t^\text{#1}}

\newcommand{\xMSSDA}{xMoSSDA}

\newcommand{\best}[1]{\textbf{#1}}
\newcommand{\sndbest}[1]{\underline{#1}}

\begin{document}
%
\title{Cross-modal Learning for Domain Adaptation \\in 3D Semantic Segmentation}
%
%
%
%

\author{Maximilian~Jaritz,
        Tuan-Hung~Vu,
        Raoul~de~Charette,
        \'Emilie Wirbel,
        and~Patrick~P\'erez
\IEEEcompsocitemizethanks{
\IEEEcompsocthanksitem Maximilian Jaritz is with Inria, Valeo and Valeo.ai.
\IEEEcompsocthanksitem Tuan-Hung Vu and Patrick P\'erez are with Valeo.ai.
\IEEEcompsocthanksitem Raoul de Charette is with Inria.
\IEEEcompsocthanksitem \'Emilie Wirbel is with Valeo and Valeo.ai.}
\thanks{}}

%
%

\markboth{}%
{Cross-modal Learning for Domain Adaptation in 3D Semantic Segmentation}
%



\IEEEtitleabstractindextext{%
\begin{abstract}
Domain adaptation is an important task to enable learning when labels are scarce. While most works focus only on the image modality, there are many important multi-modal datasets. In order to leverage multi-modality for domain adaptation, we propose cross-modal learning, where we enforce consistency between the predictions of two modalities via mutual mimicking. We constrain our network to make correct predictions on labeled data and consistent predictions across modalities on unlabeled target-domain data. Experiments in unsupervised and semi-supervised domain adaptation settings prove the effectiveness of this novel domain adaptation strategy. Specifically, we evaluate on the task of 3D semantic segmentation from either the 2D image, the 3D point cloud or from both. We leverage recent driving datasets to produce a wide variety of domain adaptation scenarios including changes in scene layout, lighting, sensor setup and weather, as well as the synthetic-to-real setup. Our method significantly improves over previous uni-modal adaptation baselines on all adaption scenarios. Our code is publicly available at this url: \url{https://github.com/valeoai/xmuda_journal}

\end{abstract}

}

\maketitle

\IEEEdisplaynontitleabstractindextext

%
\IEEEpeerreviewmaketitle

\IEEEraisesectionheading{\section{Introduction}\label{sec:introduction}}
\IEEEPARstart{S}{cene} understanding is central to many applications and, among other tasks, semantic segmentation from images has been extensively studied. However, for applications that involve interaction with the world, \textit{e.g.}, in robotics, autonomous driving or virtual reality, scenes should be understood in 3D. In this context, 3D semantic segmentation is gaining attention and an increasing number of datasets provide jointly annotated 3D point clouds (PCs) and 2D images. The modalities are complementary since PCs provide geometry while images capture texture and color.

Manual segmentation is tedious in images~\cite{cordts2016cityscapes}, but even more so in 3D PCs, because the annotator has to inspect the scene from different viewpoints~\cite{behley2019iccv}. This results in a high annotation cost. Unfortunately, the question whether sufficient ground truth can be obtained to train a large neural network can make or break a computer vision system. 
Our goal in this work is to alleviate this problem with transfer learning, in particular domain adaptation (DA), \textit{i.e.} we leverage multiple modalities (2D/3D) to improve the adaptation of a model to a target domain.

\begin{figure}
	\centering
	\includegraphics[width=0.99\linewidth]{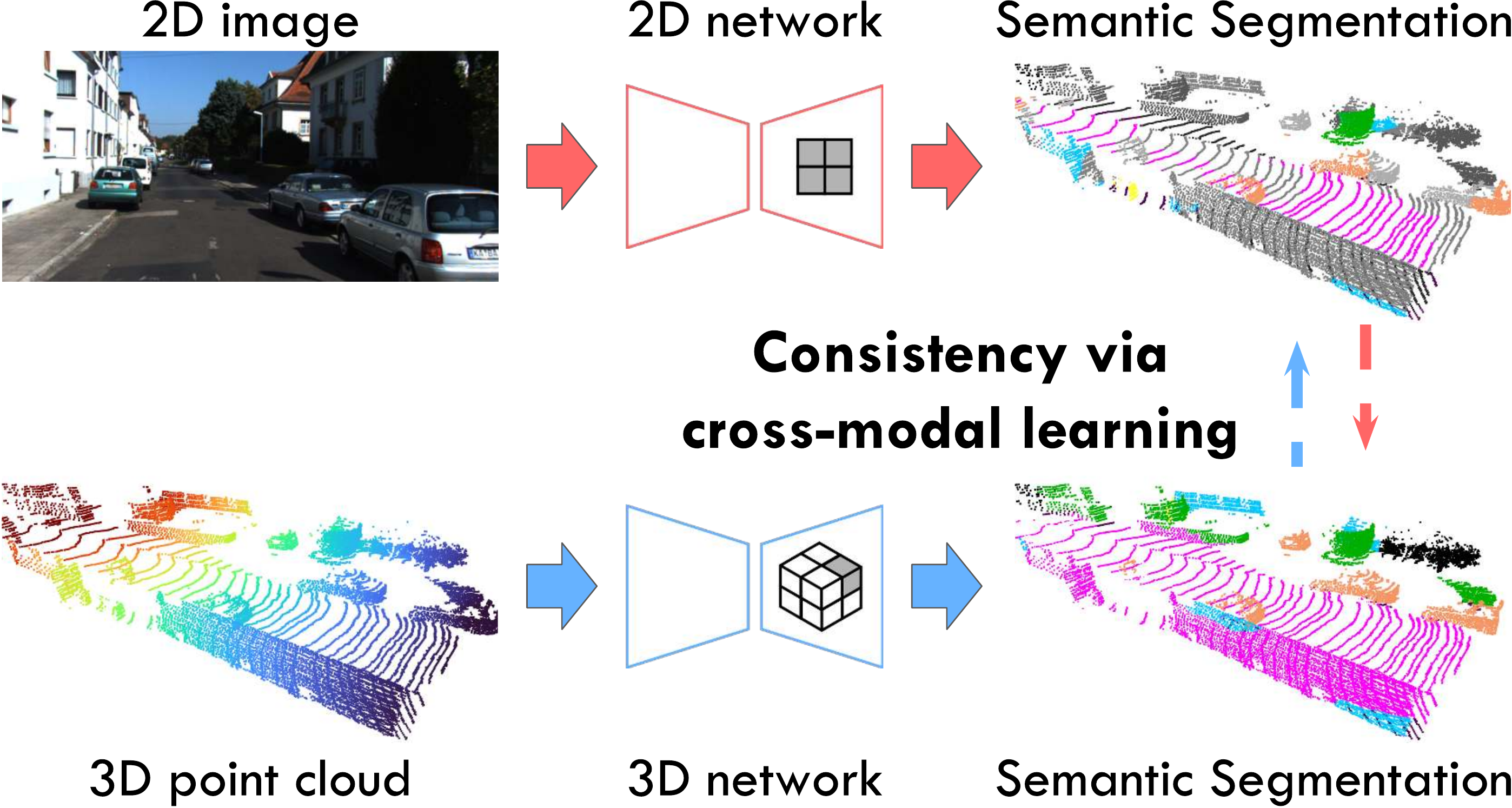}
	\caption{\textbf{Overview of the proposed cross-modal learning for domain adaptation}. Here, a 2D and a 3D network take an image and a point cloud as inputs respectively and predict their own 3D segmentation labels. Note, that the 2D predictions are uplifted to 3D. The proposed cross-modal learning enforces consistency between the 2D and 3D predictions via mutual mimicking, which proves beneficial in both unsupervised and semi-supervised domain adaptation.}
	\label{fig:teaser}
\end{figure}

We consider both unsupervised and semi-supervised DA, that is when labels are available in the source domain, but not (or only partially) in the target domain.
Most of DA literature investigates the image modality~\cite{hoffman2018cycada,hoffman-arxiv2016,vu2019advent,tsai2018learning,li2019bidirectional}, but only a few address the point-cloud modality~\cite{wu2019squeezesegv2}.
Different from these, we perform DA on images and point clouds \textit{simultaneously} with the aim to explicitly exploit multi-modality for the DA goal.

We use driving data from synchronized cameras and LiDARs, and want to profit from the fact that the domain gaps differ across these sensors. For example, a LiDAR is more robust to lighting changes (\textit{e.g.}, day/night) than a camera. On the other hand, LiDAR sensing density varies with the sensor setup while cameras always output dense images.
Our work takes advantage of the cross-modal discrepancies while preserving the best performance of each sensor thanks to the dual-head architecture -- thus avoiding that the limitations of one modality negatively affect the other modality's performance.

We propose a cross-modal loss which enforces consistency between multi-modal predictions, as depicted in Fig.~\ref{fig:teaser}. Our specifically designed dual-head architecture enables robust training by decoupling the supervised main segmentation loss from the unsupervised cross-modal loss. 

We demonstrate that our cross-modal framework proposal can be either applied in the unsupervised setting (coined xMUDA), or semi-supervised setting (coined \xMSSDA{}). 

This paper is an extension of our work~\cite{jaritz2020xmuda} which covered only UDA evaluated on three scenarios. Besides the significant expansion of the experimental evaluation (Sec.~\ref{sec:experiments}) -- including the addition of two new DA scenarios (see Fig.~\ref{fig:datasets}), the evaluation on the newly released nuScenes-Lidarseg~\cite{nuscenes2019}, and the inclusion of new baselines --, we also add a completely new use case of semi-supervised DA (SSDA) in Secs.~\ref{sec:xmssda} and \ref{sec:experiments-xmssda}. 
The original \hyperlink{https://github.com/valeoai/xmuda}{code base} of~\cite{jaritz2020xmuda} will be extended with new experiments and the SSDA set up.

In summary our contributions are:
\begin{itemize}
    \item We introduce new domain adaptation scenarios (4 unsupervised and 4 semi-supervised), for the task of 3D semantic segmentation, leveraging recent 2D-3D driving datasets with cameras and LiDARs;
    \item We propose a new DA approach with an unsupervised cross-modal loss which enforces multi-modal consistency and is complementary to other existing unsupervised techniques~\cite{lee2013pseudo};
    \item We design a robust dual-head architecture which uncouples the cross-modal loss from the main segmentation objective;
    \item We evaluate xMUDA and \xMSSDA{}, our unsupervised and semi-supervised DA scenarios respectively, and demonstrate their superior performance.
\end{itemize}

\begin{figure*}
	\centering
	\includegraphics[width=0.91\textwidth]{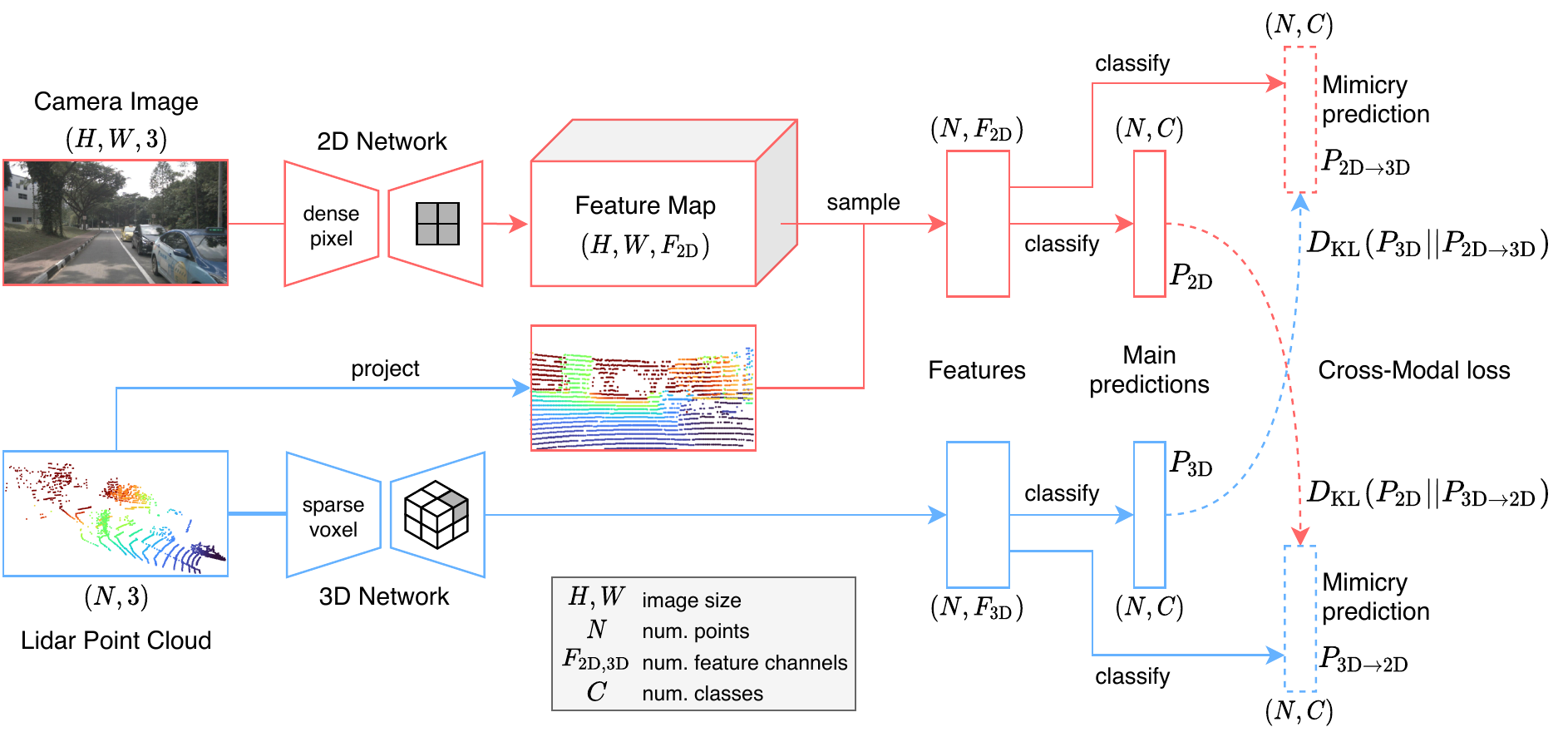}
	\caption{\textbf{Our architecture for cross-modal unsupervised learning for domain adaptation}. There are two independent network streams: a 2D~stream (in red) which takes an image as input and uses a U-Net-style 2D ConvNet~\cite{ronneberger2015unet}, as well as a 3D~stream (in blue) which takes a point cloud as input and uses a U-Net-Style 3D \mbox{SparseConvNet}~\cite{SparseConvNet}. The size of the first dimension of feature output tensors of both streams is $N$, equal to the number of 3D points. To achieve this equality, we project the 3D points, where labels exist, into the image and sample the 2D features at the corresponding pixel locations. The four segmentation outputs consist of the main predictions $P_\text{2D}, P_\text{3D}$ and the mimicry predictions $P_{\text{2D} \to \text{3D}}, P_{\text{3D} \to \text{2D}}$. We transfer knowledge across modalities using KL divergences $D_{\text{KL}}(P_\text{3D} \| P_{\text{2D} \to \text{3D}})$, where the objective of the 2D mimicry prediction is to estimate the main 3D prediction, and, vice versa, $D_{\text{KL}}(P_\text{2D} \| P_{\text{3D} \to \text{2D}})$.}
	\label{fig:architecture}
\end{figure*}

\section{Related Work}

\subsection{Unsupervised Domain Adaptation}
The past few years have seen an increasing interest in unsupervised domain adaptation (UDA) for complex perception tasks like object detection and semantic segmentation.
Under the hood of such methods lies the same spirit of learning domain-invariant representations, \textit{i.e.}, features coming from different domains should introduce insignificant discrepancies.
Some works promote adversarial training to minimize the source-target distribution shift, either on pixel-~\cite{hoffman2018cycada}, feature-~\cite{hoffman-arxiv2016} or output-space~\cite{tsai2018learning,vu2019advent}. Instead of adversarial learning, Fourier transform can also be used to stylize the source images as target~\cite{yang2020fda}.
Revisited from semi-supervised learning~\cite{lee2013pseudo}, self-training with pseudo-labels has also recently proved effective for UDA~\cite{li2019bidirectional,zou2019confidence,saporta2020esl}.

Recent works start addressing UDA in the 3D world, \textit{i.e.}, for point clouds.
LiDAR DA works are surveyed in~\cite{triess2021survey}. 
PointDAN~\cite{qin2019pointdan} proposes to jointly align local and global features used for classification.
Achituve \textit{et al.}~\cite{achituve2020self} improve UDA performance using self-supervised learning.
Wu~\textit{et al.}~\cite{wu2019squeezesegv2} adopt activation correlation alignment~\cite{morerio2017minimal} for UDA in 3D segmentation from LiDAR point clouds.
Langer~\textit{et al.}~\cite{langer2020domain} use resampling to stylize a 64 as 32-layer LiDAR, thereby aligning source and target in input point-cloud space.
Yi~\textit{et al.}~\cite{yi2020complete} also address gaps between LiDAR sampling patterns by chaining a LiDAR-specific completion network with a LiDAR-agnostic segmentation network.
In this work, we also address domain adaptation, but from a different angle, \textit{i.e.} by aligning RGB and LiDAR in output space.

To the best of our knowledge, there are no previous UDA works in 2D/3D semantic segmentation for multi-modal scenarios.
Only some consider the extra modality, \textit{e.g.}, depth, solely available at training time on the source domain and leverage such~\emph{privileged information} to boost adaptation performance~\cite{lee2018spigan,vu2019dada}.
Otherwise, we here assume all modalities are available at train and test time on both source and target domains.

\subsection{Semi-supervised Domain Adaption}
While UDA has become an active research topic, semi-supervised domain adaptation (SSDA) has so far been little-studied despite being highly relevant in practical applications.
In SSDA, we would like to transfer knowledge from a source domain with labeled data to a target domain with \textit{partially} labeled data.

Early approaches based on SVM~\cite{cortes1995support} have addressed SSDA in image classification and object detection~\cite{donahue2013semi,yao2015semi,ao2017fast}; few has been done for deep networks.
Recently, Saito~\textit{et al.}~\cite{saito2019semi} propose an adversarial SSDA learning scheme to optimize a few-shot deep classification model with minimax entropy.
Wang~\textit{et al.}~\cite{wang2020alleviating} extend UDA techniques in 2D semantic segmentation to the SSDA setting by additionally aligning feature prototypes of labeled source and target samples.
Our work is the first to address SSDA in point cloud segmentation.

\subsection{Cross-modality learning}
In our context, we define cross-modality learning as knowledge transfer between modalities. This is different from multi-modal fusion where a single model is trained supervisedly to combine complementary inputs, such as \mbox{RGB-D}~\cite{hazirbas2016fusenet,Valada_2019} or LiDAR and RGB \cite{liang2019multi,liang2018contfuse,meyer2019sensor}.

Castrejón~\textit{et al.}~\cite{castrejon2016learning} address cross-modal scene retrieval by learning a joint high-level feature representation that is agnostic to the input modality (real image, clip art, text, etc.) through enforcement of similar statistics across modalities. Gupta~\textit{et al.}~\cite{gupta2016cross} adapt the more direct feature alignment technique of distillation~\cite{hinton2015distilling} in a cross-modal setup.

Self-supervised learning produces useful representations in the absence of labels, e.g. by forcing networks with different input modalities to predict a similar output. Sayed~\textit{et al.}~\cite{sayed2018cross} minimize the cosine distance between RGB and optical flow features. Alwassel~\textit{et al.}~\cite{alwassel2020self} use clustering to generate pseudo labels and mutually train an audio and video network. Munro~\textit{et al.}~\cite{munro2020multi} use self-supervision with temporal consistency between RGB and flow.

Similar to us, Gong~\textit{et al.}~\cite{gong2021mdalu} address UDA for segmentation with RGB and LiDAR, but focus on fusing labels from multiple partial source datasets.
Instead, we use a single source dataset and explore the tasks of UDA and SSDA.

\subsection{Point cloud segmentation}
While images are dense tensors, 3D point clouds can be represented in multiple ways, which leads to competing network families evolving in parallel.

Voxels are similar to pixels, but very memory intense in their dense representation as most of them are usually empty. 
Some 3D CNNs~\cite{riegler2017octnet,tatarchenko2017octree} rely on OctTree~\cite{meagher1982geometric} to reduce memory usage but without addressing the problem of manifold dilation.
Graham~\textit{et al.}~\cite{SparseConvNet} and similar implementation~\cite{choy20194d} address the latter by using hash tables to convolve only on active voxels. This allows for very high resolution with typically only one point per voxel.
Aside from the cubic ones, cylindrical voxels are also employed \cite{zhang2020polarnet,zhu2021cylindrical}.
Finally, sparse point-voxel convolutions~\cite{tang2020searching} can benefit from lightweight support of the high resolution point-based branch.

Point-based networks perform computation in continuous 3D space and can thus directly accept point clouds as input. PointNet++~\cite{qi2017pointnetplusplus} uses point-wise convolution, max-pooling to compute global features and local neighborhood aggregation for hierarchical learning akin to CNNs. Many improvements have been proposed in this direction, such as continuous convolutions~\cite{wang2018deepcontinuous}, deformable kernels~\cite{thomas2019kpconv} or lightweight alternatives \cite{hu2020randla}.

In this work, we select SparseConvNet~\cite{SparseConvNet}, which is top performing on ScanNet~\cite{dai2017scannet}, as our 3D network. 

\section{Cross-modal Learning for DA}

Our aim is to exploit multi-modality as a source of knowledge for unsupervised learning in domain adaptation. Therefore, we propose a cross-modal learning objective, implemented as a mutual mimicking game between modalities, that drives toward consistency across predictions from different modalities. 
Of note, while our training exploits multi-modality, the 2D/3D predictions solely rely on either 2D or 3D input respectively in our architecture, making it uni-modal at inference.
Specifically, we investigate the modalities of 2D images and 3D point clouds for the task of 3D semantic segmentation as it is a core task for machine vision.

We present the network architecture in Sec.\,\ref{sec:architecture}, our framework for cross-modal unsupervised domain adaptation, coined `xMUDA', in Sec.\,\ref{sec:xmuda} and its semi-supervised version, analogously called `\xMSSDA{}', in Sec.\,\ref{sec:xmssda}.

\subsection{Architecture}\label{sec:architecture}

Our architecture predicts point-wise segmentation labels. It consists of two independent streams which respectively take a 2D image and a 3D point cloud as inputs, and output features of size $(N, F_\text{2D})$ and $(N, F_\text{3D})$ respectively, where $N$ is the number of 3D points within the camera field of view. An overview is depicted in Fig.~\ref{fig:architecture}.
By design, the 2D and 3D streams are independent, \textit{i.e.} in each stream, point-cloud semantic predictions solely rely on the respective modality.
Such an architecture choice allows better understanding of advantages and drawbacks of each modality in particular scenarios; it also helps highlight the merit of our proposed cross-modal learning.

As network backbones, we use SparseConvNet~\cite{SparseConvNet} for 3D and a modified version of U-Net~\cite{ronneberger2015unet} for 2D. Further implementation details are provided in Sec.~\ref{sec:implementation}.

\begin{figure}
	\centering
	\newcommand\height{0.45}
	\subfloat[Single head]{
		\includegraphics[height=\height\linewidth]{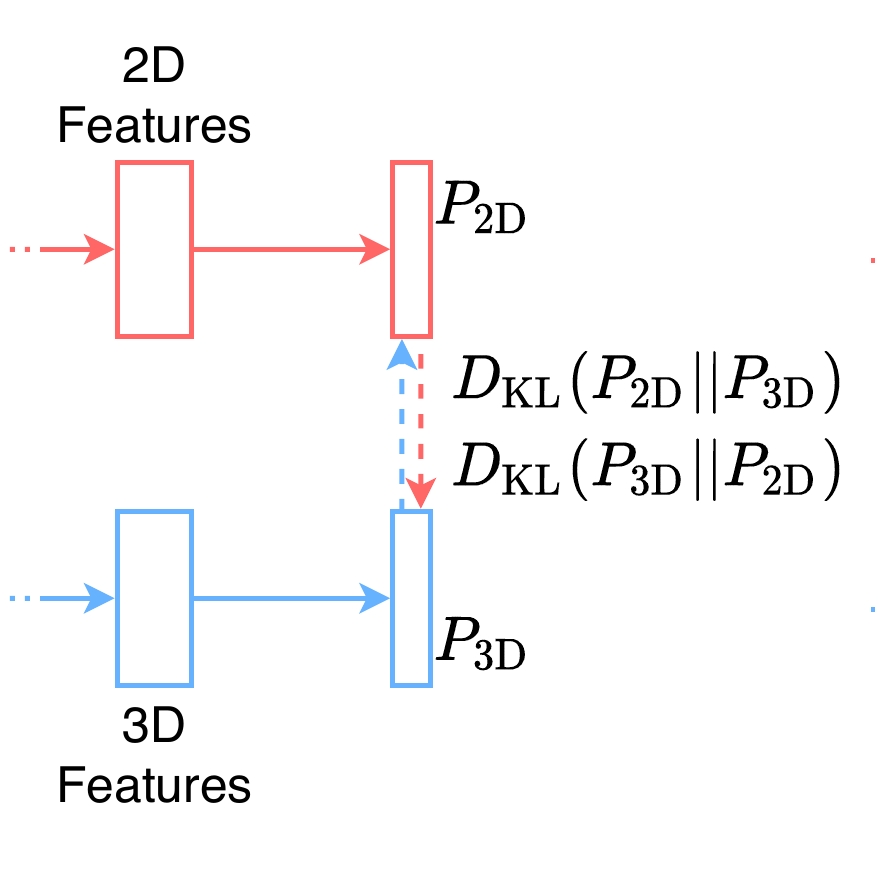}
		\label{fig:singleHead}
	}
	\subfloat[Dual head]{
		\includegraphics[height=\height\linewidth]{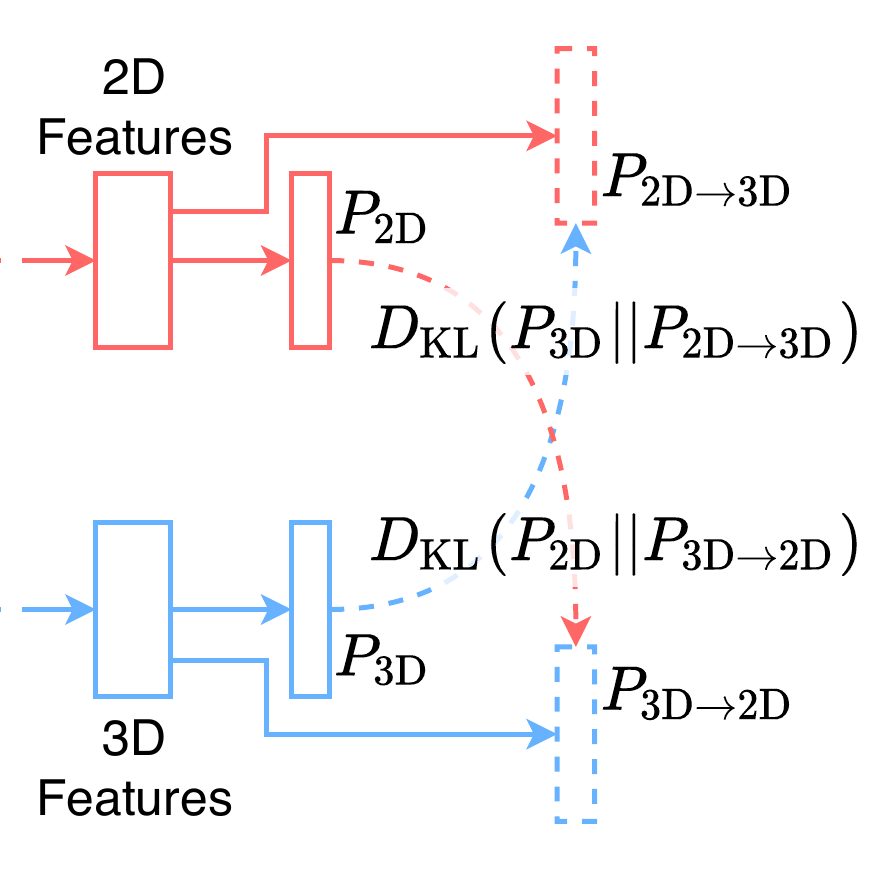}
		\label{fig:dualHead}
	}
	\caption{\textbf{Single-head vs. dual-head architecture}. \protect\subref{fig:singleHead}
	Naive way of enforcing consistency directly between main segmentation heads. \protect\subref{fig:dualHead} Our proposal of a dual-head architecture to uncouple the mimicry from the main segmentation head for more robustness. }
	\label{fig:singleVsDualHeadArch}
\end{figure}

\smallskip\noindent\textbf{Dual Segmentation Head.\,}
We call segmentation head (`classify' arrows in Fig.\,\ref{fig:architecture}) the last linear layer in the network that transforms the output features into logits followed by a softmax function to produce class probabilities.

For cross-modal learning, we establish a mimicking game between the 2D and 3D output probabilities, \textit{i.e.}, each modality should predict the other modality's output. The overall objective drives the two modalities toward an agreement, thus enforcing consistency between outputs.

In a naive approach, each modality has a single segmentation head (Fig.\,\ref{fig:singleHead}) and the cross-modal optimization objective aligns the outputs of both modalities. Unfortunately, this setup is not robust as the mimicking objective competes directly with the main segmentation objective. 
The risk is that a negative transfer from the weak modality might degrade the performance of the strong one.
This is why, in practice, one needs to down-weight the mimicry loss w.r.t. the segmentation loss to boost performance. 
However, this is a serious limitation, because down-weighting the mimicry loss also decreases its adaptation effect.

In order to address this problem, we propose to disentangle the mimicry from the main segmentation objective. Therefore, we propose a dual-head architecture as depicted in Figs.~\ref{fig:architecture} and ~\ref{fig:dualHead}. In this setup, the 2D and 3D streams both have two segmentation heads: one \textit{main} head for the best possible prediction, and one \textit{mimicry} head to estimate the other modality's output. 
The outputs of the four segmentation heads (see Fig.\,~\ref{fig:architecture}) are of size $(N, C)$, with $C$ the number of classes, such that we obtain a vector of class probabilities for each 3D point. The two main heads produce the best possible segmentation predictions, $P_\text{2D}$ and $P_\text{3D}$ respectively for each branch. The two mimicry heads estimate the other modality's output: 2D estimates 3D ($P_{\text{2D} \to \text{3D}}$) and 3D estimates 2D ($P_{\text{3D} \to \text{2D}}$).

In the following, we introduce how we use the described architecture for cross-modal learning in unsupervised (Sec.~\ref{sec:xmuda}) and semi-supervised (Sec.~\ref{sec:xmssda}) domain adaptation, respectively.

\subsection{Unsupervised Domain Adaptation (xMUDA)}\label{sec:xmuda}

\begin{figure}
	\centering
	\newcommand\width{0.99}
	\subfloat[Proposed UDA training setup]{
		\includegraphics[width=\width\linewidth]{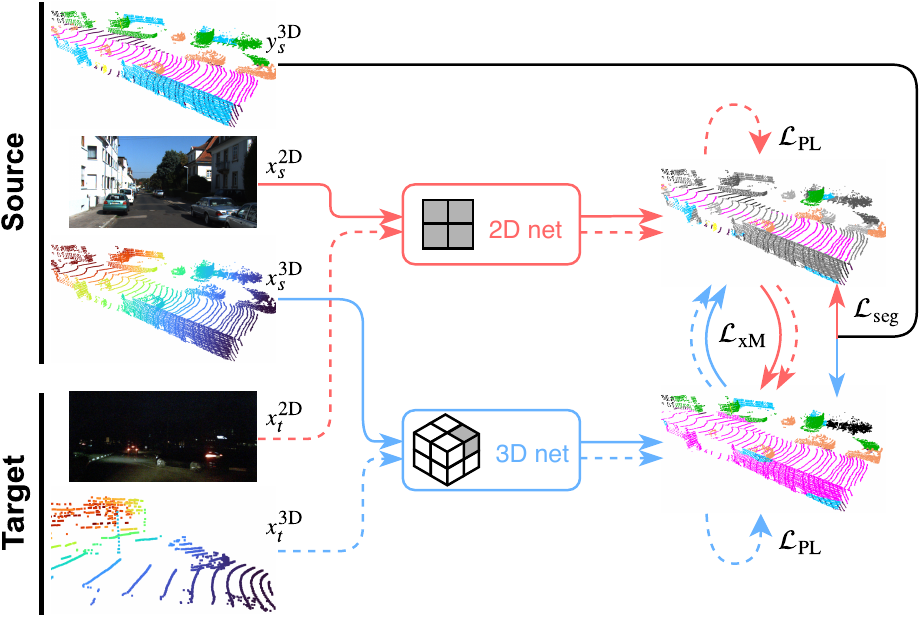}
		\label{fig:udaTrainingSetup}
	}\\
	\subfloat[UDA on multi-modal data]{
		\includegraphics[width=\width\linewidth]{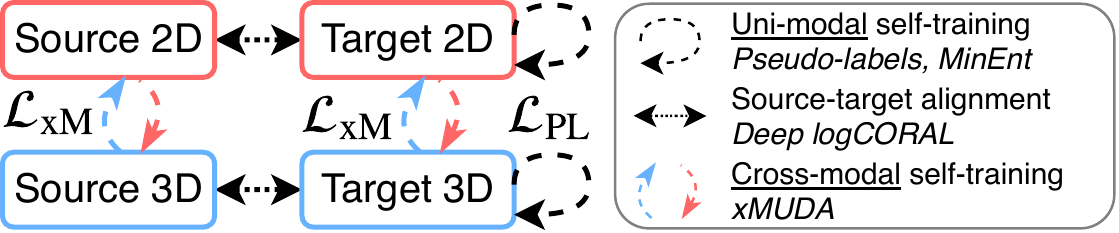}
		\label{fig:dataLossesOverview}
	}
	\caption{\textbf{Cross-modal training with adaptation}. \protect\subref{fig:udaTrainingSetup} xMUDA learns from supervision on the source domain (plain lines) and self-supervision on the target domain (dashed lines) thanks to cross-modal learning between 2D/3D. 
	\protect\subref{fig:dataLossesOverview} We consider four data subsets: Source 2D, Target 2D, Source 3D and Target 3D. In contrast to existing techniques, xMUDA introduces a cross-modal self-training mechanism for UDA.
	}
	\label{fig:uda}
\end{figure}

We propose xMUDA, cross-modal unsupervised domain adaptation, which considers a \textit{source-domain} dataset $\mathcal{S}$, where each sample consists of a 2D image $\xs{2D} \in \mathbb{R}^{H \times W \times 3}$, a 3D point cloud $\xs{3D} \in \mathbb{R}^{N \times 3}$ and 3D segmentation labels $\ys \in \llbracket 1, C \rrbracket^{N}$ with $C$ classes, as well as a \textit{target-domain} dataset~$\mathcal{T}$, lacking annotations, where each sample only consists of an image $\xt{2D}$ and a point cloud $\xt{3D}$.

In the following, we define the usual supervised learning setup, our cross-modal loss $\mathcal{L}_{\text{xM}}$, and an additional variant `xMUDA\textsubscript{PL}' that further uses pseudo-labels to boost performance. An overview of the learning setup is in Fig.~\ref{fig:udaTrainingSetup}. The difference between our \textit{cross-modal} learning and existing \textit{uni-modal} UDA techniques, such as as Pseudo-labels~\cite{lee2013pseudo}, MinEnt~\cite{vu2019advent} or Deep logCORAL~\cite{morerio2017minimal} is visualized in Fig.~\ref{fig:dataLossesOverview}.

\subsubsection{Supervised Learning}

The main goal of 3D segmentation is learned through cross-entropy in a classical supervised fashion on the source-domain data. 
Denoting $\bm{P}_{\bm{x}}\in[0,1]^{N\times C}$ the soft-classification map associated by the segmentation model to the $N$ 3D points of interest, for a given input $\bm{x}$, the segmentation loss $\mathcal{L}_\text{seg}$ of each network stream (2D and 3D) for a given training sample in $\mathcal{S}$ reads: 
\begin{align}\label{eq:SegLoss}
\mathcal{L}_{\text{seg}}(\bm{x}, \bm{y}^\text{3D}) = - \frac{1}{N}\sum_{n=1}^N \sum_{c=1}^C \bm{y}^{(n, c)} \log \bm{P}_{\bm{x}}^{(n, c)},
\end{align}
where $\bm{x}$ is either $\xs{2D}$ or $\xs{3D}$ and $\bm{y}^\text{3D}$ equals $\ys$. We denote tensor entries' indices as superscript.

\subsubsection{Cross-Modal Learning}

The goal of unsupervised learning across modalities is twofold. Firstly, we want to transfer knowledge from one modality to the other on the target-domain dataset; for example, if one modality is more sensitive than the other to the domain shift, 
then the robust modality should teach the sensitive one the correct class in the target domain where no labels are available. Secondly, we want to design an auxiliary objective on source and target domains, where the task is to estimate the other modality's prediction. By mimicking not only the class with maximum probability, but the whole distribution like in teacher-student distillation~\cite{hinton2015distilling}, more information is exchanged, leading to softer labels.

We choose the KL divergence for the cross-modal loss $\mathcal{L}_\text{xM}$ and define it as follows:
\begin{align}\label{eq:xmloss}
\mathcal{L}_{\text{xM}}(\bm{x}) &= \bm{D}_\text{KL}(\bm{P}_{\bm{x}}^{(n, c)} \| \bm{Q}_{\bm{x}}^{(n, c)})\\
&= - \frac{1}{N} \sum_{n=1}^N \sum_{c=1}^C \bm{P}_{\bm{x}}^{(n, c)} \log \frac{\bm{P}_{\bm{x}}^{(n, c)}}{\bm{Q}_{\bm{x}}^{(n, c)}},
\end{align}
with $(\bm{P}, \bm{Q}) \in  \{(\bm{P}_\text{2D}, P_{\text{3D} \to \text{2D}}), (\bm{P}_\text{3D}, P_{\text{2D} \to \text{3D}})\}$ where $\bm{P}$ is the target distribution from the main prediction which is to be estimated by the mimicking prediction $\bm{Q}$.
This loss is applied on the source and the target domain as it does not require ground-truth labels and is the key to our proposed domain adaptation framework. In the source domain, $\mathcal{L}_\text{xM}$ can be seen as an auxiliary mimicry loss in addition to the main segmentation loss $\mathcal{L}_\text{seg}$.

The final objective for each network stream (2D and 3D) is the combination of the segmentation loss $\mathcal{L}_\text{seg}$ on source-domain data and the cross-modal loss $\mathcal{L}_\text{xM}$ on both domains:
\begin{multline}\label{eq:completeObjective}
\min_{\theta}\Big[\frac{1}{|\mathcal{S}|}\sum_{\bm{x}_s\in\mathcal{S}} \Big(\mathcal{L}_\text{seg}(\bm{x}_s, \ys) + \lambda_s \mathcal{L}_\text{xM}(\bm{x}_s)\Big) \\
 + \frac{1}{|\mathcal{T}|}\sum_{\bm{x}_t\in\mathcal{T}} \lambda_t \mathcal{L}_\text{xM}(\bm{x}_t)\Big],
\end{multline}
where $\lambda_s, \lambda_t$ are hyperparameters to weight $\mathcal{L}_\text{xM}$ on source and target domain respectively and $\theta$ are the network weights of either the 2D or the 3D stream.

There are parallels between our approach and Deep Mutual Learning~\cite{zhang2018deep} in training two networks in collaboration and using the KL divergence as mimicry loss. However, unlike this work, our cross-modal learning establishes consistency across modalities (2D/3D) \textit{without} supervision.

\subsubsection{Self-training with Pseudo-Labels}

Cross-modal learning is complementary to pseudo-labeling~\cite{lee2013pseudo} used originally in semi-supervised learning and recently in UDA ~\cite{li2019bidirectional,zou2019confidence}.
To benefit from both, once having optimized a model with Eq.\,~\ref{eq:completeObjective}, we extract pseudo-labels offline, selecting highly-confident labels based on the predicted class probability.
Then, we train again from scratch using the produced pseudo-labels for an additional segmentation loss on the target-domain training set.
The optimization problem writes:
\begin{multline}\label{eq:completeObjectiveWithPL}
\min_{\theta}\Big[\frac{1}{|\mathcal{S}|}\sum_{\bm{x}_s} \Big(\mathcal{L}_\text{seg}(\bm{x}_s, \ys) + \lambda_s \mathcal{L}_\text{xM}(\bm{x}_s)\Big) \\
+ \frac{1}{|\mathcal{T}|}\sum_{\bm{x}_t} \Big(\lambda_\text{PL} \mathcal{L}_\text{seg}(\bm{x}_t, \hat{\bm{y}}_t^\text{3D}) + \lambda_t \mathcal{L}_\text{xM}(\bm{x}_t)\Big)\Big],
\end{multline}
where $\lambda_\text{PL}$ weights the pseudo-label segmentation loss and $\hat{\bm{y}}^\text{3D}$ are the pseudo-labels. For clarity, we will refer to the xMUDA variant that uses additional self-training with pseudo-labels as xMUDA\textsubscript{PL}.

\subsection{Semi-supervised Domain Adaptation (\xMSSDA{})}\label{sec:xmssda}

Cross-modal learning can also be used in semi-supervised domain adaptation, thus benefiting from a small portion of labeled data in the target domain.

Formally, we consider in \xMSSDA{} a labeled source-domain dataset $\mathcal{S}$ where each sample contains an image $\xs{2D}$, a point cloud $\xs{3D}$ and labels $\ys$. Different from unsupervised learning, the target-domain set consists of a usually small labeled part~$\mathcal{T}_{\ell}$ where each sample holds an image $\bm{x}_{t\ell}^\text{2D}$, a point cloud $\bm{x}_{t\ell}^\text{3D}$ and labels $\bm{y}^\text{3D}_{t\ell}$, as well as an, often larger, unlabeled part~$\mathcal{T}_u$ where each sample consists only of an image $\bm{x}_{tu}^\text{2D}$ and a point cloud $\bm{x}_{tu}^\text{3D}$.

\subsubsection{Supervised Learning}
Unlike xMUDA, we do not only apply the segmentation loss $\mathcal{L}_{\text{seg}}(\bm{x}, \bm{y}^\text{3D})$ of Eq.~\ref{eq:SegLoss} on the source-domain dataset~$\mathcal{S}$, but also on the labeled target-domain dataset~$\mathcal{T}_{\ell}$: The segmentation loss in Eq.~\ref{eq:SegLoss} thus applies both to samples $(\bm{x}, \bm{y}^\text{3D}) \in \{ (\xs{2D}, \ys), (\xs{3D}, \ys) \}$ in~$\mathcal{S}$ and to samples $(\bm{x}, \bm{y}^\text{3D}) \in \{ (\bm{x}_{t\ell}^\text{2D}, \bm{y}^\text{3D}_{t\ell}), (\bm{x}_{t\ell}^\text{3D}, \bm{y}^\text{3D}_{t\ell}) \}$ in~$\mathcal{T}_{\ell}$. 
Note that, in practice, we train on source and target domains at the same time by concatenating examples from both in a batch.

\subsubsection{Cross-Modal Learning}

We apply the unsupervised cross-modal loss $\mathcal{L}_\text{xM}$ of Eq.~\ref{eq:xmloss} on all datasets, \textit{i.e.}, (labeled) source-domain $\mathcal{S}$, labeled target-domain~$\mathcal{T}_{\ell}$, and unlabeled target-domain dataset~$\mathcal{T}_u$. The latter,~$\mathcal{T}_u$, is a typically large portion of unlabeled data compared to a usually much smaller labeled portion~$\mathcal{T}_{\ell}$. Subsequently, it is beneficial to also exploit~$\mathcal{T}_u$ with an unsupervised loss, such as cross-modal learning. The complete objective is a combination of supervised segmentation loss $\mathcal{L}_{\text{seg}}$ where labels are available (\textit{i.e.}, in sets $\mathcal{S}$ and~$\mathcal{T}_{\ell}$), and unsupervised cross-modal loss $\mathcal{L}_{\text{xM}}$ everywhere (\textit{i.e.}, on $\mathcal{S}, \mathcal{T}_{\ell}$ and~$\mathcal{T}_u$) which enforces consistency between the 2D and 3D predictions. It writes:
\begin{equation}
\begin{split}\label{eq:completeObjectiveXmssda}
\min_{\theta}\Big[ & \frac{1}{|\mathcal{S}|}\sum_{\bm{x}_s\in\mathcal{S}} \Big(\mathcal{L}_\text{seg}(\bm{x}_s, \ys) + \lambda_s \mathcal{L}_\text{xM}(\bm{x}_s)\Big) \\
+ & \frac{1}{|\mathcal{T}_{\ell}|}\sum_{\bm{x}_{t\ell}\in\mathcal{T}_{\ell}} \Big(\mathcal{L}_\text{seg}(\bm{x}_{t\ell}, \bm{y}^\text{3D}_{t\ell}) + \lambda_{t\ell} \mathcal{L}_\text{xM}(\bm{x}_{t\ell})\Big) \\
+ & \frac{1}{|\mathcal{T}_u|}\sum_{\bm{x}_{tu}\in\mathcal{T}_u} \lambda_{tu} \mathcal{L}_\text{xM}(\bm{x}_{tu})\Big],
\end{split}
\end{equation}
where $\lambda_s$, $\lambda_{t\ell}$ and $\lambda_{tu}$ are the weighting hyperparameters for $\mathcal{L}_\text{xM}$. In practice we choose $\lambda_s = \lambda_{t\ell}$ for simplicity.

\subsubsection{Self-training with Pseudo-Labels}

As in the unsupervised setting, we extend semi-supervised cross-modal learning to also benefit from pseudo-labels. 
After having trained a model with Eq.~\ref{eq:completeObjectiveXmssda}, we use the model to generate predictions on the unlabeled target-domain dataset~$\mathcal{T}_u$ and extract highly-confident pseudo-labels which are used to train again from scratch with the following objective:
\begin{equation}
\begin{split}\label{eq:completeObjectiveXmssdaWithPL}
\min_{\theta}\Big[ & \frac{1}{|\mathcal{S}|}\sum_{\bm{x}_s\in\mathcal{S}} \Big(\mathcal{L}_\text{seg}(\bm{x}_s, \ys) + \lambda_s \mathcal{L}_\text{xM}(\bm{x}_s)\Big) \\
+ & \frac{1}{|\mathcal{T}_{\ell}|}\sum_{\bm{x}_{t\ell}\in\mathcal{T}_{\ell}} \Big(\mathcal{L}_\text{seg}(\bm{x}_{t\ell}, \bm{y}^\text{3D}_{t\ell}) + \lambda_{t\ell} \mathcal{L}_\text{xM}(\bm{x}_{t\ell})\Big) \\
+ & \frac{1}{|\mathcal{T}_u|}\sum_{\bm{x}_{tu}\in\mathcal{T}_u} \Big( \lambda_\text{PL} \mathcal{L}_\text{seg}(\bm{x}_{tu}, \hat{\bm{y}}_{tu}^\text{3D}) + \lambda_{tu} \mathcal{L}_\text{xM}(\bm{x}_{tu})\Big) \Big],
\end{split}
\end{equation}
where $\lambda_\text{PL}$ is weighting the pseudo-label segmentation loss and $\hat{\bm{y}}^\text{3D}$ are pseudo-labels. We call this variant \xMSSDA{}\textsubscript{PL}.

\section{Experiments}\label{sec:experiments}
For evaluation, we identified five domain adaptation~(DA) scenarios relevant to autonomous driving, shown in Fig.~\ref{fig:datasets}, and evaluated our proposals against recent baselines.

Hereafter, we first describe the datasets~(Sec.~\ref{sec:datasets}), the implementation backbone and training details~(Sec.~\ref{sec:implementation}), and then evaluate xMUDA~(Sec.~\ref{sec:experiments-xmuda}) and \xMSSDA{}~(Sec.~\ref{sec:experiments-xmssda}). Finally, we extend our cross-modal framework to fusion~(Sec.~\ref{sec:extFusion}), demonstrating its global benefit.

\subsection{Datasets}\label{sec:datasets}
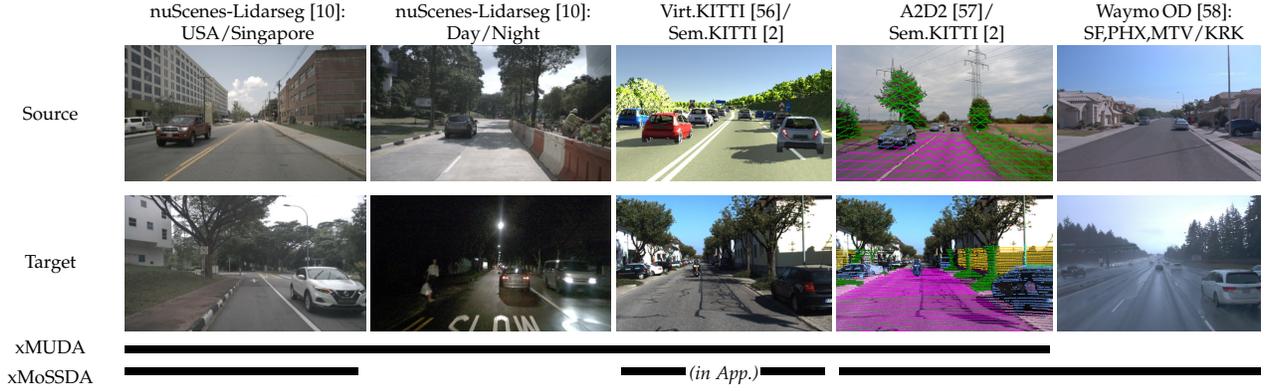
\begin{figure*}
	\centering
	\scriptsize
	\newcommand\height{1.8cm}
	\begin{tabular}{cc@{\hspace{0pt}}c@{\hspace{0pt}}c@{\hspace{0pt}}c@{\hspace{0pt}}c}
		& \makecell[c]{nuScenes-Lidarseg\cite{nuscenes2019}: \\ USA/Singapore } &
		\makecell[c]{nuScenes-Lidarseg\cite{nuscenes2019}: \\ Day/Night } &
		\makecell[c]{Virt.KITTI\cite{gaidon2016virtual}/ \\ Sem.KITTI\cite{behley2019iccv}} & 
		\makecell[c]{A2D2\cite{aev2019}/ \\ Sem.KITTI\cite{behley2019iccv}} &
		\makecell[c]{Waymo\,OD\cite{sun2020scalability}: \\ SF,PHX,MTV/KRK} \\
		\adjustbox{valign=c}{Source} &
		\adjustbox{valign=c}{\includegraphics[height=\height]{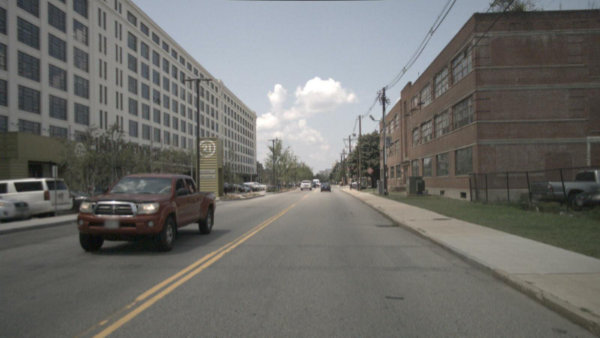} } &
		\adjustbox{valign=c}{\includegraphics[height=\height]{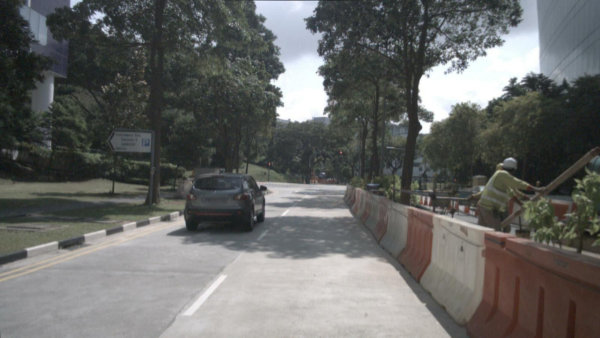} } &
		\adjustbox{valign=c}{\includegraphics[height=\height]{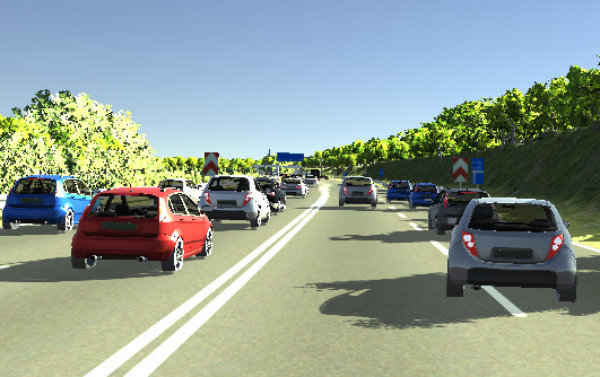} } &
		\adjustbox{valign=c}{\includegraphics[height=\height]{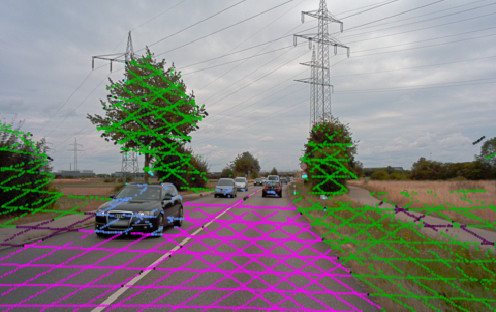} } &
		\adjustbox{valign=c}{\includegraphics[height=\height]{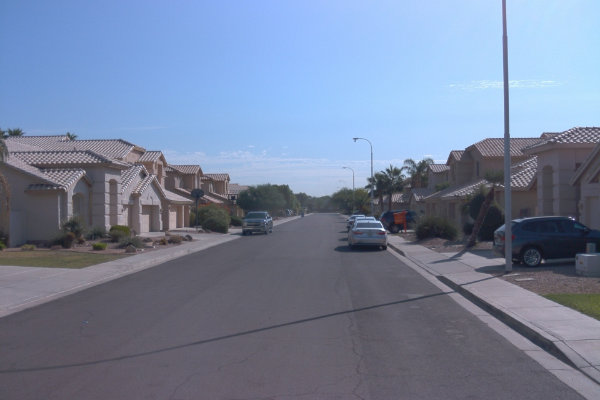} } 
		\vspace{0.2cm}\\
		Target &
		\adjustbox{valign=c}{\includegraphics[height=\height]{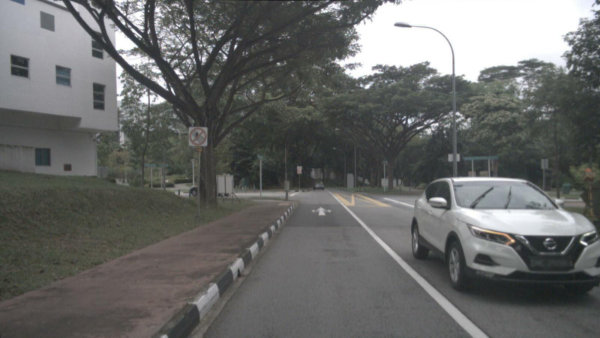} }&
		\adjustbox{valign=c}{\includegraphics[height=\height]{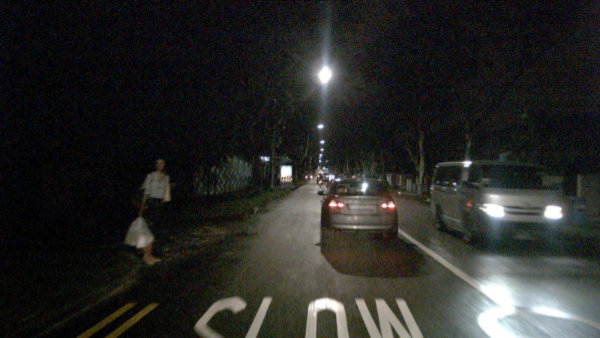} } &
		\adjustbox{valign=c}{\includegraphics[height=\height]{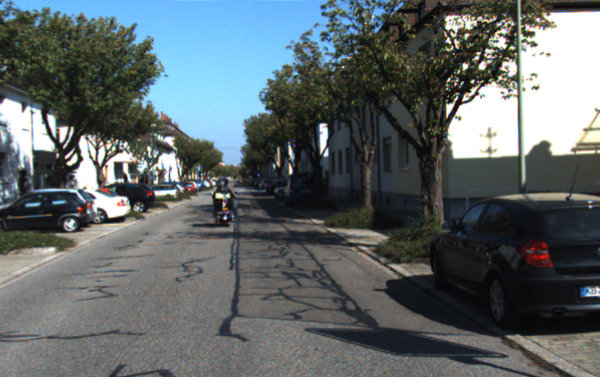} } &
		\adjustbox{valign=c}{\includegraphics[height=\height]{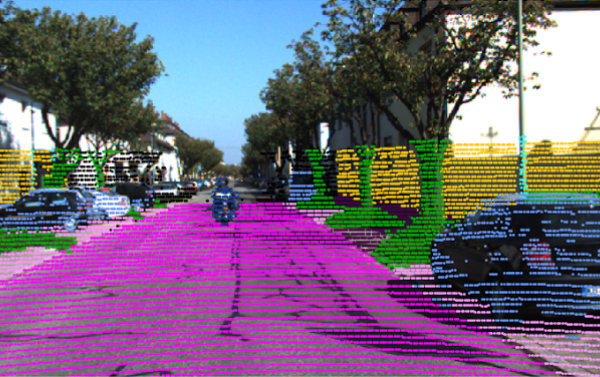} } &
		\adjustbox{valign=c}{\includegraphics[height=\height]{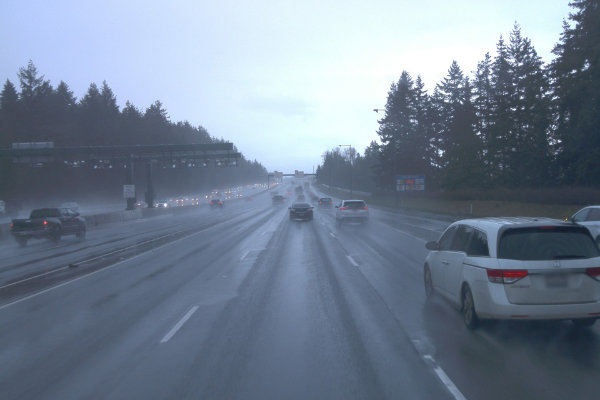} }
		\vspace{0.1cm} \\
		xMUDA & \multicolumn{5}{l}{
			\begin{tikzpicture}
				\fill[fill=black,black] (0.1,0) rectangle (12.4, 0.1);
			\end{tikzpicture}
		}
		\\
        \makecell[c]{\vspace{0.8em}\xMSSDA{}}
        & 
        \multicolumn{5}{l}{
            \begin{tikzpicture}
    			\fill[fill=black,black] (0,0) rectangle (3.1, 0.1);
    			\fill[fill=black,black] (6.6,0) rectangle (7.45, 0.1)
    			node[right] at (7.41, 0.0) {\textit{(in App.)}};
    			\fill[fill=black,black] (8.45,0) rectangle (9.3, 0.1);
    			\fill[fill=black,black] (9.5,0) rectangle (15.1, 0.1);
    		\end{tikzpicture}
		}
		\\
		\vspace{-0.8em}
	\end{tabular}
	\vspace{-0.4cm}
	\caption{\textbf{Overview of the five proposed DA scenarios}. We generate the nuScenes-Lidarseg~\cite{nuscenes2019} splits using metadata. The third and fourth DA scenarios use both SemanticKITTI~\cite{behley2019iccv} as target-domain dataset and either the synthetic VirtualKITTI~\cite{gaidon2016virtual} or the real A2D2 dataset~\cite{aev2019} as source-domain dataset. Note that we show the A2D2/SemanticKITTI scenario with LiDAR overlay to visualize the density difference and resulting domain gap. Last, Waymo\,OD~\cite{sun2020scalability} features a source-domain dataset in the cities of San Francisco (SF), Phoenix (PHX) and Mountain View (MTV) and a target-domain dataset in Kirkland (KRK). We evaluate xMUDA on scenarios 1-4 and \xMSSDA{} on scenarios 1, 4, 5 and 3 in Appendix~\ref{sec:appendix:noGainFromSour}}
	\label{fig:datasets}
\end{figure*}

To compose our domain adaptation scenarios displayed in Fig.~\ref{fig:datasets}, we leveraged public datasets nuScenes-Lidarseg~\cite{nuscenes2019}, VirtualKITTI~\cite{gaidon2016virtual}, SemanticKITTI~\cite{behley2019iccv}, A2D2~\cite{aev2019} and Waymo Open Dataset (Waymo\,OD)~\cite{sun2020scalability}. The split details are in Tab.~\ref{tab:splits}.
Our scenarios cover typical DA challenges like change in scene layout, between right and left-hand-side driving in the \textit{nuScenes-Lidarseg:\,USA/Singapore} scenario, lighting changes, between day and night in \textit{nuScenes-Lidarseg:\,Day/Night}, synthetic-to-real data, between simulated depth and RGB to real LiDAR and camera in \textit{VirtualKITTI/SemanticKITTI}, different sensor setups and characteristics like resolution/FoV in \textit{A2D2/SemanticKITTI} and weather changes between sunny San Francisco, Phoenix, Mountain View and rainy Kirkland in \textit{Waymo\,OD: SF,PHX,MTV/KRK}.

In all datasets, the LiDAR and the camera are synchronized and calibrated, allowing 2D/3D projections. For consistency across datasets, we only use the front camera's images, even when multiple cameras are available.

Waymo\,OD does not provide point-wise 3D segmentation labels so we leverage 3D object bounding-box labels. Points lying inside a box are labeled as that class and points outside of all boxes are labeled as background. 

To compensate for source/target classes mismatch (\textit{e.g.}, VirtualKITTI/SemanticKITTI) or accommodate for classes scarcity, we apply a custom class mapping detailed in Appendix \ref{sec:appendix:datasetSplits}. Note that VirtualKITTI provides depth maps so we simulate LiDAR scanning via uniform point sampling.

All training data and splits can be reproduced with our code, and more details are in Appendix \ref{sec:appendix:datasetSplits}.

\begin{table}
\centering
\resizebox{.45\textwidth}{!}{%
    \begin{tabular}{llrp{0.01cm}rr}
        \toprule
        & & Source~$\mathcal{S}$ & \multicolumn{3}{c}{Target~$\mathcal{T}$} \\
        \cmidrule(r){3-3}
        \cmidrule(r){4-6}
        & Scenario & Train & & Train & Val/Test \\
        \midrule
        \multirow{4}{*}{ \rotatebox[origin=c]{90}{UDA} } & nuSc-Lidarseg:\,USA/Singap. & 15,695 &  & 9,665 & 2,770/2,929 \\
        & nuSc-Lidarseg:\,Day/Night & 24,745 &  & 2,779 & 606/602 \\
        & Virt.KITTI/Sem.KITTI & 2,126 &  & 18,029 & 1,101/4,071 \\
        & A2D2/Sem.KITTI & 27,695 &  & 18,029 & 1,101/4,071 \\
        \midrule
        \multirow{4}{*}{ \rotatebox[origin=c]{90}{SSDA} } & nuSc-Lidarseg:\,USA/Singap. & 15,695 & \makecell[l]{
        $\mathcal{T}_\ell$ \\
        $\mathcal{T}_u$} & \makecell[r]{
        2,884 \\
        6,781 } & 2,770/2,929 \\
        & Virt.KITTI/Sem.KITTI & 2,126 & \makecell[l]{
        $\mathcal{T}_\ell$ \\
        $\mathcal{T}_u$} & \makecell[r]{
        5,642 \\
        32,738 } & 1,101/4,071 \\
        & A2D2/Sem.KITTI & 27,695 & \makecell[l]{
        $\mathcal{T}_\ell$ \\
        $\mathcal{T}_u$} & \makecell[r]{
        5,642 \\
        32,738 } & 1,101/4,071 \\
        & \makecell[l]{Waymo\,OD:\\ SF,PHX,MTV/KRK} & 158,081 & \makecell[l]{
        $\mathcal{T}_\ell$ \\
        $\mathcal{T}_u$} & \makecell[r]{
        11,853 \\
        94,624 } & 3,943/3,932 \\
        \bottomrule
    \end{tabular}}
	\caption{\textbf{Size of the splits in frames for all proposed DA scenarios}. While there is a single target-domain training set in UDA ($\mathcal{T}$), there are two in SSDA: a labeled target-domain training set~$\mathcal{T}_{\ell}$ and a (much larger) unlabeled set~$\mathcal{T}_u$.}
	\label{tab:splits}
\end{table}

\subsection{Implementation Details}
\label{sec:implementation}

In the following, we briefly introduce our implementation. Please refer to our code for further details.

\textbf{2D Network.}
We use a modified version of U-Net~\cite{ronneberger2015unet} with ResNet34~\cite{he2016resnet} encoder and a decoder with transposed convolutions and skip connections.
To lift the 2D features to 3D, we subsample the output feature map of size $(H, W, F_\text{2D})$ at the pixel locations where the $N$ 3D points project. Hence, the 2D network takes an image $\bm{x}^\text{2D}$ as input and outputs features of size $(N, F_\text{2D})$.

\textbf{3D Network.}
We use the official SparseConvNet~\cite{SparseConvNet} implementation and a U-Net architecture with 6 times downsampling. The voxel size is set to 5cm which is small enough to only have one 3D point per voxel. Thus, the 3D network takes a point cloud $\bm{x}^\text{3D}$ as input and outputs features of size $(N, F_\text{3D})$.

\textbf{Training.}
We employ standard 2D/3D data augmentation and log-smoothed class weights to address class-imbalance.
In PyTorch, to compute the KL divergence for the cross-modal loss, we \textit{detach} the target variable to only backpropagate in either the 2D or the 3D network.
We train with a batch size of 8, Adam optimizer with $\beta_1=0.9, \beta_2=0.999$, and train 30k iterations for the scenario with the small VirtualKITTI dataset and 100k iterations for all other scenarios.
At each iteration we compute and accumulate gradients on the source and target batch, jointly training the 2D and 3D stream. 
To fit the training into a single GPU with 11GB of memory, we resize the images and additionally crop them in VirtualKITTI and SemanticKITTI.

For the pseudo-label variants, xMUDA\textsubscript{PL} and \xMSSDA{}\textsubscript{PL}, we generate the pseudo-labels offline as in~\cite{li2019bidirectional} with trained models xMUDA and \xMSSDA{}, respectively. Then, we retrain from scratch, additionally using the pseudo-labels, optimizing Eqs.~\ref{eq:completeObjectiveWithPL} and \ref{eq:completeObjectiveXmssdaWithPL}, respectively. Importantly, we only use the last checkpoint to generate the pseudo-labels -- as opposed to using the best weights which would provide a supervised signal.

\subsection{xMUDA}\label{sec:experimentsXmuda}
\label{sec:experiments-xmuda}

\begin{table*}
\centering
\resizebox{0.90\textwidth}{!}{%
	\begin{threeparttable}
        \begin{tabular}{lcccccccccccc}
            \toprule
            & \multicolumn{3}{c}{nuSc-Lidarseg:\,USA/Singap.} & \multicolumn{3}{c}{nuSc-Lidarseg:\,Day/Night} & \multicolumn{3}{c}{Virt.KITTI/Sem.KITTI} & \multicolumn{3}{c}{A2D2/Sem.KITTI} \\
            \cmidrule(r){2-4}
            \cmidrule(r){5-7}
            \cmidrule(r){8-10}
            \cmidrule(r){11-13}
            Method & 2D & 3D & 2D+3D & 2D & 3D & 2D+3D & 2D & 3D & 2D+3D & 2D & 3D & 2D+3D \\
            \midrule
            Baseline (src only) & 58.4 & 62.8 & 68.2 & 47.8 & 68.8 & 63.3 & 26.8 & 42.0 & 42.2 & 34.2 & 35.9 & 40.4 \\
            \midrule
            Deep logCORAL~\cite{morerio2017minimal} & \sndbest{64.4} & 63.2 & 69.4 & 47.7 & 68.7 & 63.7 & 41.4\tnote{*} & 36.8 & 47.0 & 35.1\tnote{*} & 41.0 & 42.2 \\
            MinEnt~\cite{vu2019advent} & 57.6 & 61.5 & 66.0 & 47.1 & 68.8 & 63.6 & 39.2 & 43.3 & 47.1 & 37.8 & 39.6 & 42.6 \\
            PL~\cite{li2019bidirectional} & 62.0 & \sndbest{64.8} & \sndbest{70.4} & 47.0 & \textbf{69.6} & 63.0 & 21.5 & 44.3 & 35.6 & 34.7 & 41.7 & \sndbest{45.2} \\
            FDA~\cite{yang2020fda} & 60.8 & - & - & 48.4 & - & - & 32.8\tnote{*} & - & - & 37.6\tnote{*} & - & - \\
            \midrule
            xMUDA & \sndbest{64.4} & 63.2 & 69.4 & \sndbest{55.5} & \sndbest{69.2} & \textbf{67.4} & \sndbest{42.1} & \sndbest{46.7} & \sndbest{48.2} & \sndbest{38.3} & \sndbest{46.0} & 44.0 \\
            xMUDA\textsubscript{PL} & \textbf{67.0} & \textbf{65.4} & \textbf{71.2} & \textbf{57.6} & \textbf{69.6} & \sndbest{64.4} & \textbf{45.8} & \textbf{51.4} & \textbf{52.0} & \textbf{41.2} & \textbf{49.8} & \textbf{47.5} \\
            \midrule
            \midrule
            Oracle & 75.4 & 76.0 & 79.6 & 61.5 & 69.8 & 69.2 & 66.3 & 78.4 & 80.1 & 59.3 & 71.9 & 73.6 \\
            \midrule
            Domain gap (O-B) & 17.0 & 13.3 & 11.5 & 13.6 & 1.1 & 5.9 & 39.5 & 36.4 & 37.9 & 25.1 & 36.0 & 33.2 \\
            \bottomrule
        \end{tabular}
    	\begin{tablenotes}
    		\item[*] The 2D network is trained with batch size 6 instead of 8 to fit into GPU memory.
    	\end{tablenotes}
	\end{threeparttable}}
	\caption{\textbf{xMUDA experiments on 3D semantic segmentation}. We report the mIoU result (with \best{best} and \sndbest{2nd best}) on the target set for each network stream (2D and 3D) as well as the ensembling result taking the mean of the 2D and 3D probabilities (`2D+3D'). We provide the lower bound `Baseline (src only)' which is trained on the source set $\mathcal{S}$, but not on the target set~$\mathcal{T}$, as well as the upper bound `Oracle' which is trained supervisedly on the target set~$\mathcal{T}$ using labels. We further indicate the `Domain gap' which is the difference between the Oracle and Baseline score. `Deep logCORAL'~\cite{morerio2017minimal}, `MinEnt'~\cite{vu2019advent} and `PL'~\cite{li2019bidirectional} are 2D/3D uni-modal UDA baselines, whereas `FDA` is 2D-only. The two variants `xMUDA' and `xMUDA\textsubscript{PL}' are our methods. We evaluate on four UDA scenarios (see Fig.\,\protect{\ref{fig:datasets}}). For the nuScenes-Lidarseg dataset~\cite{nuscenes2019} (`nuSc-Lidarseg'), we generate the splits with different locations (USA/Singapore) and different time (Day/Night). VirtualKITTI~\cite{gaidon2016virtual} (`Virt.KITTI') to SemanticKITTI~\cite{behley2019iccv} explores challenging synthetic-to-real adaptation. The domain gap between the two real datasets A2D2~\cite{aev2019}/SemanticKITTI~\cite{behley2019iccv} (`Sem.KITTI') lies mainly in sensor resolution. }
	\label{tab:resultsXmuda}
\end{table*}

We evaluate xMUDA on four unsupervised domain adaptation scenarios and compare against uni-modal UDA methods: Deep logCORAL~\cite{morerio2017minimal}, entropy minimization (MinEnt)~\cite{vu2019advent}, pseudo-labeling (PL)~\cite{li2019bidirectional} and Fourier domain adaptation (FDA)~\cite{yang2020fda}.
For~\cite{li2019bidirectional} the image-2-image translation part was excluded due to its instability, high training complexity and incompatibility with LiDAR data.
Regarding the three other uni-modal techniques, we adapt the published implementations to our settings.
For all, we searched for the best respective hyperparameters.
For the 2D-only baseline FDA~\cite{yang2020fda}, we implement the full MTB method, \textit{i.e.} with entropy, ensembling of three models to generate pseudo-labels and re-training.
We found no 3D-only UDA baseline directly applicable to our scenarios. Instead, we compare against LiDAR transfer~\cite{langer2020domain} -- largely outperformed on their own scenario -- in Appendix~\ref{sec:appendix:lidarTransfer}.

We report mean Intersection over Union (mIoU) of the target test set for 3D segmentation in Tab.\,~\ref{tab:resultsXmuda}. We evaluate on the test set using the checkpoint that achieved the best score on the validation set.
In addition to the scores of the 2D and 3D model, we show the ensembling result (`2D+3D') which is obtained by taking the mean of the predicted 2D and 3D probabilities after softmax.
The uni-modal UDA baselines~\cite{li2019bidirectional,morerio2017minimal,vu2019advent} are applied separately on each modality, and FDA~\cite{yang2020fda} is a 2D-only UDA baseline.

Furthermore, we provide the results of a lower bound, `Baseline (src only)', which is not domain adaptation as it is only trained on the source-domain dataset and an upper bound, `Oracle', trained only on target \textit{with} labels\footnote{Except for the Day/Night oracle, where we use 50\%/50\% source/target batches to prevent overfitting due to the small target size.}. We also indicate the `Domain gap (O-B)', computed as the difference between Oracle and Baseline. It shows that the intra-dataset domain gaps (nuScenes-Lidarseg:\,USA/Singapore, Day/Night) , in~$[1.1, 17.0]$, are much smaller than the inter-dataset domain gaps (A2D2/SemanticKITTI, VirtualKITTI/SemanticKITTI), in~$[25.1, 39.5]$. It suggests that a change in sensor setup (A2D2/SemanticKITTI) is actually a very hard domain adaptation problem, similar to the synthetic-to-real case (VirtualKITTI/SemanticKITTI). Importantly, note that the scores are not comparable between A2D2/SemanticKITTI and VirtualKITTI/SemanticKITTI, as they use a different number of classes, 10 and 6 respectively.

xMUDA --using the cross-modal loss but \textit{not} PL-- brings a significant adaptation effect on all four UDA scenarios compared to `Baseline' and most often outperforms all uni-modal UDA baselines.
xMUDA\textsubscript{PL} achieves the best score everywhere with the only exception of Day/Night~2D+3D where xMUDA is better. Further, cross-modal learning and self-training with pseudo-labels (PL) are complementary as their combination in xMUDA\textsubscript{PL} typically yields a higher score, up to $+4,7$, than each separate technique.
The 2D/3D oracle scores indicate that overall LiDAR (3D) is always the strongest modality, which resonates with the choice of 3D segmentation task.
However, xMUDA consistently improves \textit{both} modalities (2D and 3D) \textit{i.e.}, even the strong modality can learn from the weaker one. 
A notable example for 3D is at Night where xMUDA~(69.2) outperforms `Baseline'~(68.8), despite a narrow Domain gap~(1.1) related to the active LiDAR sensing capability.
The dual-head architecture might be key here: each modality can improve its main segmentation head independently from the other modality, because the consistency is achieved indirectly through the mimicking heads.

We also observe a regularization effect thanks to xMUDA. For example on VirtualKITTI/SemanticKITTI, the methods `Baseline' and `PL' perform very poorly on the 2D modality (26.8 and 21.5) due to overfitting on the very small VirtualKITTI dataset, while 3D is more stable (42.0 and 44.3). In contrast, xMUDA performs better as 3D can regularize 2D. This regularization even enables the benefit of pseudo-labels, because xMUDA\textsubscript{PL} achieves an even better score.

\begin{figure*}
	\centering
	\scriptsize
	\includegraphics[width=1\textwidth]{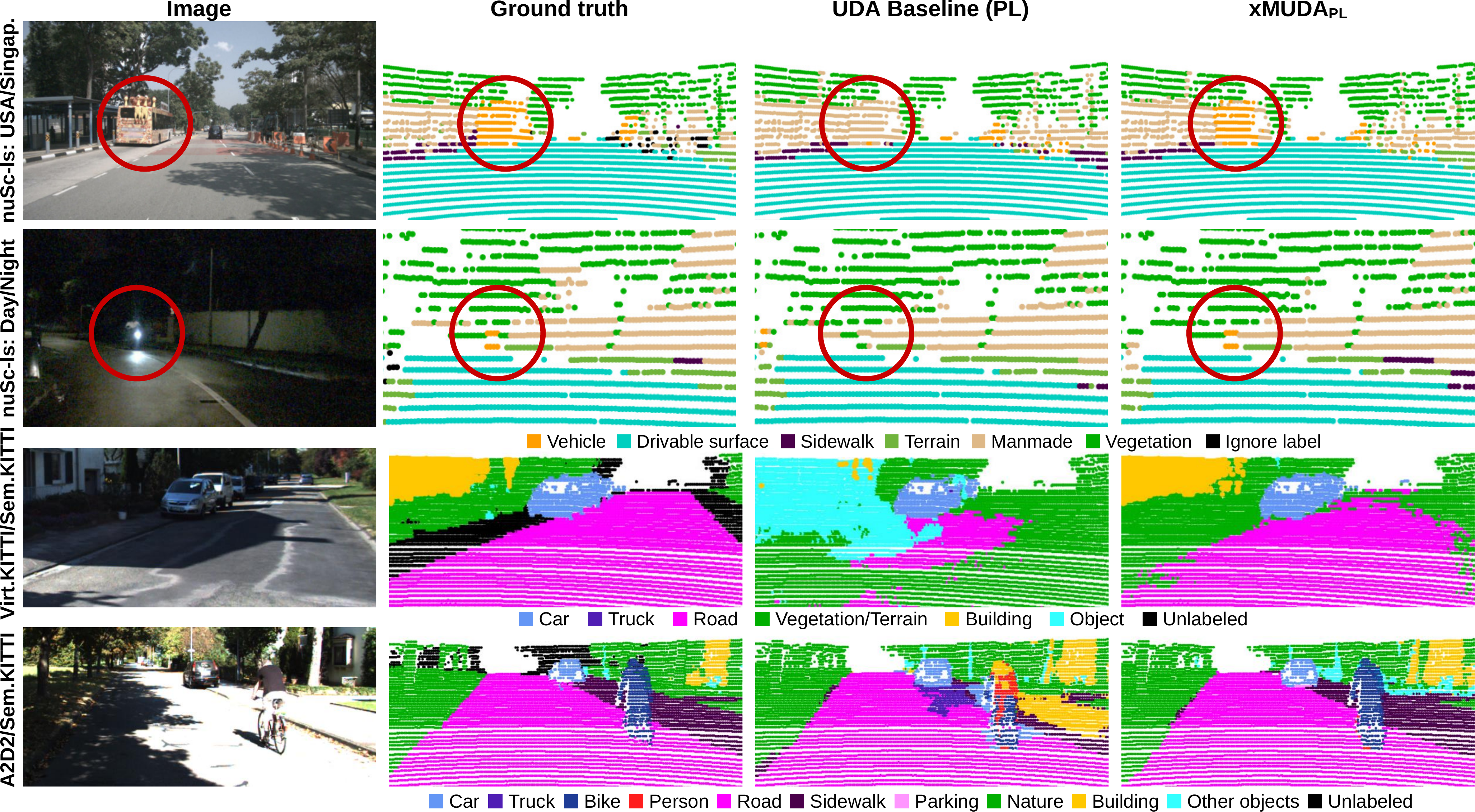}
	\caption{\textbf{Qualitative results for xMUDA}. We show the ensembling result (2D+3D) on the target test set for UDA Baseline (PL) and xMUDA\textsubscript{PL}.\\
	--\,\textbf{nuScenes-Lidarseg:\,USA/Singapore}: UDA Baseline (PL) fails to correctly classify the bus while xMUDA\textsubscript{PL} succeeds.\\
	--\,\textbf{nuScenes-Lidarseg:\,Day/Night}: A motorcycle in oncoming traffic. The visual appearance is very different during the day (motorcycle visible) than during the night (only the headlight visible). The uni-modal UDA baseline is not able to learn this new appearance. However, if information between camera and robust-at-night LiDAR is exchanged in xMUDA\textsubscript{PL}, it is possible to detect the motorcycle correctly at night.\\
	--\,\textbf{A2D2/SemanticKITTI}:\,xMUDA\textsubscript{PL} helps to stabilize and increase segmentation performance when there are sensor changes (3x16-layer LiDAR with different angles to 64-layer LiDAR).\\
	--\,\textbf{VirtualKITTI/SemanticKITTI}:\,The UDA baseline (PL) poorly segments the building and road while xMUDA\textsubscript{PL} succeeds.}
	\label{fig:qualitativeResultsXmuda}
\end{figure*}

Qualitative results are presented in Fig.\,~\ref{fig:qualitativeResultsXmuda}, showing the versatility of xMUDA across all proposed UDA scenarios. 
Here, the benefit of xMUDA\textsubscript{PL} over uni-modal PL baseline is evident in inter-dataset scenarios (last two rows), and more subtly on the nuScenes scenarios (first two rows) looking at the vehicle class.
We provide additional qualitative results in Appendix~\ref{sec:appendix:additionalQualRes} and a video at \mbox{\url{http://tiny.cc/cross-modal-learning}}.

We also successfully experiment our method in some opposite adaptation directions, with details in Appendix~\ref{sec:appendix:oppositeDir}.

\subsection{\xMSSDA}
\label{sec:experiments-xmssda}

\begin{table*}
\centering
\resizebox{0.8\textwidth}{!}{%
	\begin{threeparttable}
        \begin{tabular}{llccccccccc}
        \toprule
        & & \multicolumn{3}{c}{\makecell[l] {nuSc-Lidarseg:\\ USA/Singap.}} & \multicolumn{3}{c}{\makecell[l] {\\ A2D2/Sem.KITTI}} & \multicolumn{3}{c}{\makecell[l] {Waymo\,OD:\\SF,PHX,MTV/KRK}} \\
        \cmidrule(r){3-5}
        \cmidrule(r){6-8}
        \cmidrule(r){9-11}
        Method & Train set & 2D & 3D & 2D+3D & 2D & 3D & 2D+3D & 2D & 3D & 2D+3D \\
        \midrule
        Baseline (src only) & $\mathcal{S}$ & 58.8 & 63.2 & 68.5 & 37.9 & 32.8 & 43.3 & 61.4 & 50.8 & 64.4 \\
        Baseline (lab. trg only) & $\mathcal{T}_\ell$ & 70.5 & 74.1 & 74.2 & 51.3 & 57.7 & 59.2 & 56.5 & 57.1 & 60.3 \\
        Baseline (src and lab. trg) & $\mathcal{S}$~+~$\mathcal{T}_\ell$ & 72.3 & 73.1 & 78.1 & 54.8 & 62.4 & 66.2 & 64.5 & 56.3 & 69.3 \\
        \midrule\midrule
        Domain gap ($\mathcal{S}$ vs. $\mathcal{S}$ + $\mathcal{T}_{\ell}$) & & 13.5 & 9.9 & 9.6 & 16.9 & 29.5 & 22.9 & 3.2 & 5.5 & 4.9 \\
        \midrule
        xMUDA & $\mathcal{S}$~+~$\mathcal{T}_u$ & 63.1 & 64.2 & 67.8 & 38.6 & 44.5 & 44.4 & 61.8 & 54.0 & 66.7 \\
        xMUDA\textsubscript{PL} & $\mathcal{S}$~+~$\mathcal{T}_u$ & 66.2 & 65.1 & 70.1 & 41.4 & 49.5 & 48.6 & \sndbest{68.3} & 55.2 & \sndbest{71.9} \\
        \midrule
        Deep logCORAL~\cite{morerio2017minimal} & $\mathcal{S}$~+~$\mathcal{T}_\ell$~+~$\mathcal{T}_u$ & 71.7 & 73.1 & 78.2 & 55.1\tnote{*} & 62.2 & 64.7\tnote{*} & 61.4 & 56.5 & 66.1 \\
        MinEnt~\cite{vu2019advent} & $\mathcal{S}$~+~$\mathcal{T}_\ell$~+~$\mathcal{T}_u$ & 72.6 & 73.3 & 76.6 & 56.3 & 62.5 & 65.0 & 64.3 & 56.6 & 69.1 \\
        PL~\cite{li2019bidirectional} & $\mathcal{S}$~+~$\mathcal{T}_\ell$~+~$\mathcal{T}_u$ & 73.6 & \sndbest{74.4} & \textbf{79.3} & \sndbest{57.2} & \sndbest{66.9} & \sndbest{68.5} & 67.4 & 56.7 & 70.2 \\
        \midrule
        \xMSSDA{} & $\mathcal{S}$~+~$\mathcal{T}_\ell$~+~$\mathcal{T}_u$ & \sndbest{74.3} & 74.1 & 78.5 & 56.5 & 63.4 & 65.9 & 65.2 & \sndbest{57.4} & 69.4 \\
        \xMSSDA{}\textsubscript{PL} & $\mathcal{S}$~+~$\mathcal{T}_\ell$~+~$\mathcal{T}_u$ & \textbf{75.5} & \textbf{74.8} & \sndbest{78.8} & \textbf{59.1} & \textbf{68.2} & \textbf{70.7} & \textbf{70.1} & \textbf{58.5} & \textbf{73.1} \\
        \midrule\midrule
        Unsupervised advantage & & 3.1 & 1.7 & 0.7 & 4.3 & 5.8 & 4.5 & 5.6 & 2.2 & 3.8 \\
        \tiny{(relative)} & & \tiny{(+4.3\%)} & \tiny{(+2.3\%)} & \tiny{(+0.9\%)} & \tiny{(+7.8\%)} & \tiny{(+9.3\%)} & \tiny{(+6.8\%)} & \tiny{(+8.7\%)} & \tiny{(+3.9\%)} & \tiny{(+5.5\%)} \\
        \bottomrule
        \end{tabular}
	\begin{tablenotes}
	\item[*] The 2D network is trained with batch size 6 instead of 8 to fit into GPU memory.
	\end{tablenotes}
	\end{threeparttable}}
	\caption{\textbf{\xMSSDA{} experiments on 3D semantic segmentation}. We report the mIoU (with \best{best} and \sndbest{2nd best}) on the target set for each network stream (2D and 3D) as well as the ensembling result taking the mean of the 2D and 3D probabilities (2D+3D). In semi-supervised adaptation (SSDA), we have a source dataset $\mathcal{S}$ like in UDA, while, unlike UDA, the target dataset has a small labeled part~$\mathcal{T}_{\ell}$ and a large unlabeled part~$\mathcal{T}_u$. We provide three baselines where we train either on source only ($\mathcal{S}$), on labeled target only ($\mathcal{T}_{\ell}$) or on both ($\mathcal{S}$~+~$\mathcal{T}_{\ell}$) with ratio 50\%/50\% in each batch. For comparison, we report `xMUDA' and `xMUDA\textsubscript{PL}' results that do not use $\mathcal{T}_{\ell}$. The three uni-modal SSDA baselines `Deep logCORAL'~\cite{morerio2017minimal}, `MinEnt'~\cite{vu2019advent} and `PL'~\cite{li2019bidirectional} as well as our cross-modal methods `\xMSSDA{}' and `\xMSSDA{}\textsubscript{PL}' are trained supervisedly on $\mathcal{S}$~+~$\mathcal{T}_{\ell}$ with ratio 50\%/50\% in each batch and unsupervisedly on $\mathcal{T}_u$. 
	We report the domain gap and the `Unsupervised advantage', i.e. the difference between \xMSSDA{}\textsubscript{PL} and `Baseline (src and lab. trg)' and relative improvement. 
	We evaluate on SSDA scenarios: nuScenes-Lidarseg~\cite{nuscenes2019} (USA/Singapore), A2D2~\cite{aev2019}/SemanticKITTI~\cite{behley2019iccv} and Waymo\,OD\cite{sun2020scalability}.
	}
	\label{tab:resultsXmssda}
\end{table*}

\begin{figure*}
	\centering
	\includegraphics[width=1\textwidth]{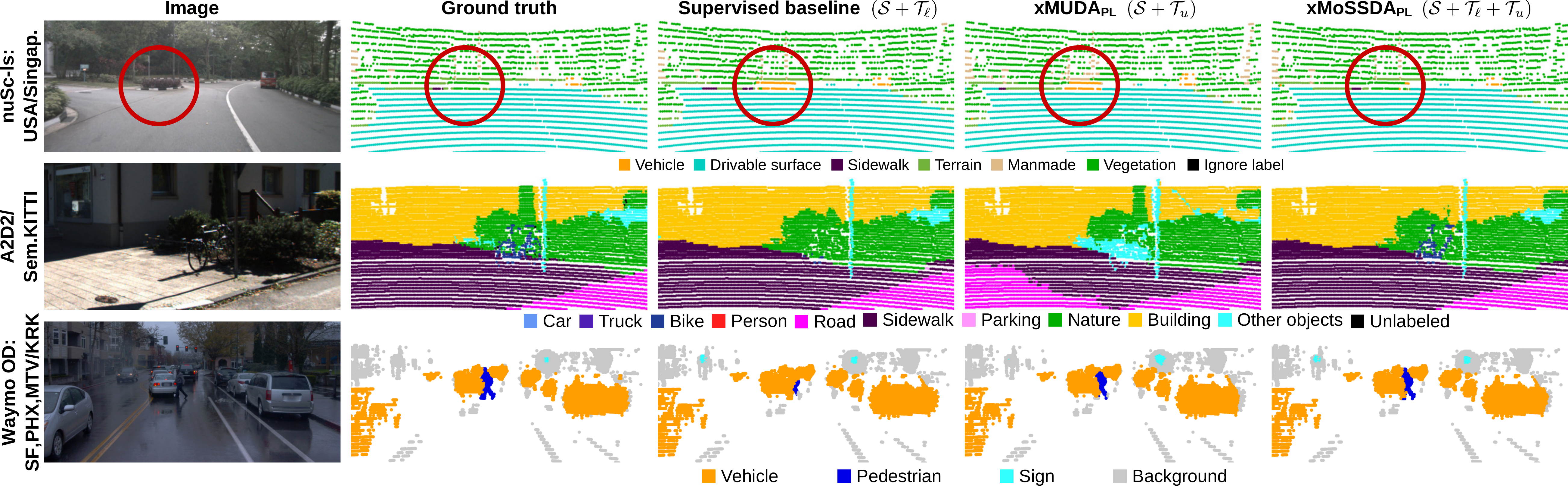}
	\caption{\textbf{Qualitative results for \xMSSDA{}}. We show the ensembling result (2D+3D) on the target test set for the supervised baseline (trained on $\mathcal{S}$~+~$\mathcal{T}_{\ell}$), xMUDA\textsubscript{PL} (trained on $\mathcal{S}$~+~$\mathcal{T}_u$) and \xMSSDA{}\textsubscript{PL} (trained on $\mathcal{S}$~+~$\mathcal{T}_{\ell}$ +~$\mathcal{T}_u$).\\
	--\,\textbf{nuScenes-Lidarseg:\,USA/Singapore}: A bush is mistakenly classified as vehicle by the supervised baseline and xMUDA\textsubscript{PL}, but correctly classified as vegetation by \xMSSDA{}\textsubscript{PL}.\\
	--\,\textbf{A2D2/SemanticKITTI}: The bike in the center is not distinguished from `Nature' background by the supervised baseline, but is so by xMUDA\textsubscript{PL}, although still wrongly classified, while \xMSSDA{}\textsubscript{PL} is correct.\\
	--\,\textbf{Waymo\,OD: SF,PHX,MTV/KRK}: Segmentation of the pedestrian with xMUDA\textsubscript{PL} is better than with the supervised baseline while it is best with \xMSSDA{}\textsubscript{PL}.}
	\label{fig:qualitativeResultsXmossda}
\end{figure*}

In this section, we evaluate \xMSSDA{} on domain adaptation scenarios nuSc-Lidarseg:\,USA/Singap., A2D2/SemanticKITTI and Waymo\,OD. To create practically relevant SSDA conditions, we make sure that the unlabeled target dataset $\mathcal{T}_u$ is much larger than the labeled target dataset $\mathcal{T}_{\ell}$. Hence, splits differ from UDA (see Tab.~\ref{tab:splits}).

We compare \xMSSDA{} against eight baselines.
Three baselines are purely supervised, either trained on source only ($\mathcal{S}$), labeled target only ($\mathcal{T}_{\ell}$) or on source \textit{and} labeled target\footnote{The latter is trained with 50\%/50\% examples from $\mathcal{S}$ and~$\mathcal{T}_{\ell}$, \textit{i.e.}, a training batch of size 8 has 4 random samples from $\mathcal{S}$ and 4 from~$\mathcal{T}_{\ell}$.} ($\mathcal{S}$~+~$\mathcal{T}_{\ell}$).
Additionally we report two UDA baselines, xMUDA and xMUDA\textsubscript{PL}, which use source and \textit{unlabeled} target ($\mathcal{S}$~+~$\mathcal{T}_u$).
Last, we report three SSDA baselines (trained on $\mathcal{S}$~+~$\mathcal{T}_{\ell}$ +~$\mathcal{T}_u$) adapted from uni-modal UDA baselines~\cite{morerio2017minimal,vu2019advent,li2019bidirectional} as follows: we train similarly to the supervised baseline on $\mathcal{S}$~+~$\mathcal{T}_{\ell}$ with 50\%/50\% batches, but add the respective domain adaptation loss on~$\mathcal{T}_u$. Our semi-supervised proposals, \xMSSDA{} and \xMSSDA{}\textsubscript{PL}, are also trained in this manner. 

To achieve $|\mathcal{T}_u| \gg |\mathcal{T}_{\ell}|$, we include \textit{unannotated} data for Waymo OD and SemanticKITTI (we use hidden test set) into $\mathcal{T}_u$. Hence, it is impossible to train an Oracle like in Tab.~\ref{tab:resultsXmuda}.
Instead, we answer the question: ``How much can we improve over the supervised baseline by additionally training on $\mathcal{T}_u$?''. We coin it the `unsupervised advantage' computed as the difference between \xMSSDA{}\textsubscript{PL} ($\mathcal{S}$~+~$\mathcal{T}_{u}$~+~$\mathcal{T}_{\ell}$) and supervised baseline ($\mathcal{S}$~+~$\mathcal{T}_{\ell}$).
Note that we exclude nuSc-Lidarseg:\,Day/Night and Virt.KITTI/Sem.KITTI due to small target/source datasets respectively, but still evaluate Virt.KITTI/Sem.KITTI in Appendix~\ref{sec:appendix:noGainFromSour}.

We report the mIoU for 3D segmentation in Tab.~\ref{tab:resultsXmssda}. Note that the latter results cannot be compared to Tab.~\ref{tab:resultsXmuda} since splits differ.
We observe in Tab.~\ref{tab:resultsXmssda}, similar to Tab.~\ref{tab:resultsXmuda}, that the domain gap is much larger in the inter-dataset adaptation A2D2/SemanticKITTI (max. 29.5) than in intra-dataset adaptation on nuSc-Lidarseg:\,USA/Singap. (max. 13.5) and Waymo\,OD, (max. 4.5).
As expected, xMUDA and xMUDA\textsubscript{PL} ($\mathcal{S}$~+~$\mathcal{T}_{u}$) improve over the baseline ($\mathcal{S}$), but are (with the exception of Waymo OD) worse than the baseline ($\mathcal{S}$~+~$\mathcal{T}_\ell$).
We also observe that \xMSSDA{} improves 2D and 3D more (max. 2.0) than ensemble result 2D+3D (max. 0.4), w.r.t. to Baseline ($\mathcal{S}$~+~$\mathcal{T}_\ell$). We ascribe this behavior to the ensembling (2D+3D) performing best when 2D and 3D predictions differ, though our cross-modal loss aligns 2D/3D predictions.
In \xMSSDA{}\textsubscript{PL}, the separate 2D/3D pseudo-labels act as counterweight to this alignment, resulting in comparable 2D, 3D and 2D+3D improvement.

Finally, \xMSSDA{}\textsubscript{PL} outperforms all baselines by up to +2.2 in 8 out of 9 cases.
Results in Fig.~\ref{fig:qualitativeResultsXmossda} show better segmentation of thin structures (bush, bike, pedestrian).
	
\subsection{Extension to Fusion}\label{sec:extFusion}

\begin{figure}
	\centering
	\newcommand\height{0.45}
	\subfloat[Vanilla Fusion]{
		\includegraphics[height=\height\linewidth]{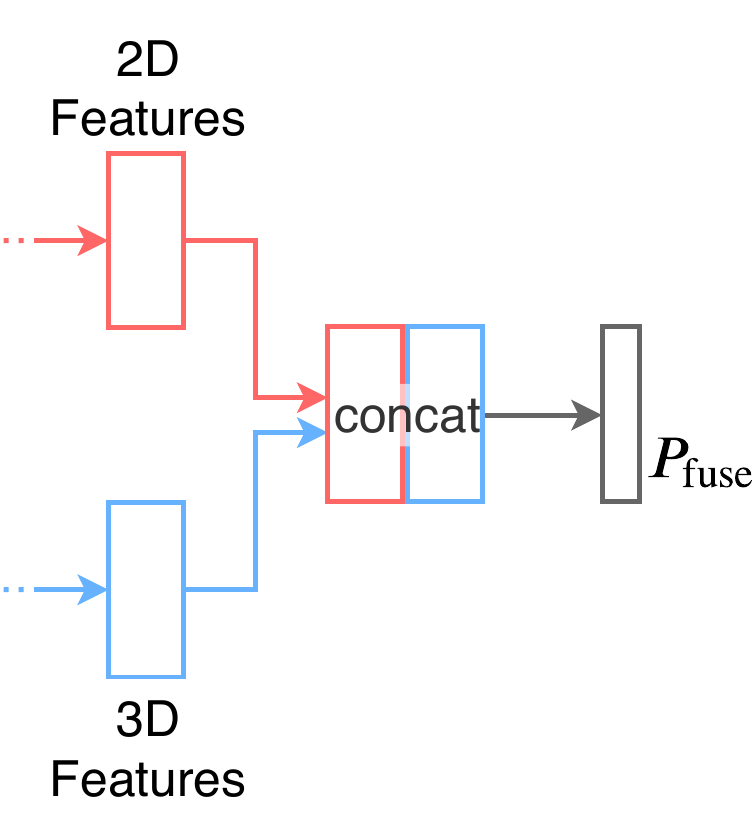}
		\label{fig:architectureFusionVanilla}
	}
	\subfloat[xMUDA Fusion]{
		\includegraphics[height=\height\linewidth]{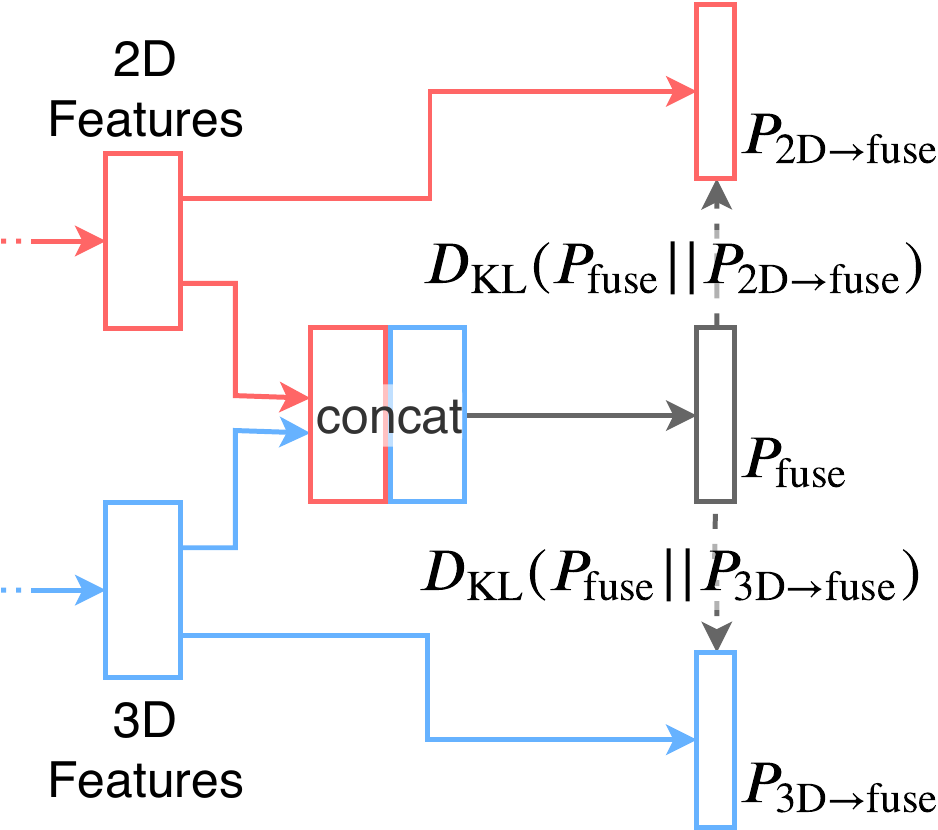}
		\label{fig:architectureFusionXmuda}
	}
	\caption{\textbf{Architectures for fusion}. \protect\subref{fig:architectureFusionVanilla} In Vanilla Fusion the 2D and 3D features are concatenated, fed into a linear layer with ReLU to mix them and followed by another linear layer and softmax to obtain a fused prediction $\boldsymbol{P}_\text{fuse}$. \protect\subref{fig:architectureFusionXmuda} In xMUDA Fusion, we add two uni-modal outputs $\boldsymbol{P}_{\text{2D} \to \text{fuse}}$ and $\boldsymbol{P}_{\text{3D} \to \text{fuse}}$ that are used to mimic the fusion output $\boldsymbol{P}_\text{fuse}$.}
	\label{fig:architectureVanillaVsAS}
\end{figure}

So far, we used an architecture with independent 2D/3D streams. However, can xMUDA also be applied in a \textit{fusion} setup where both modalities make a joint prediction?

A common fusion architecture is late fusion where the features from different sources are concatenated (see Fig.\,~\ref{fig:architectureFusionVanilla}). However, when merging the main 2D/3D branches into a unique fused head, we can no longer apply cross-modal learning (as in Fig.\,~\ref{fig:singleHead}). To address this problem, we propose `xMUDA Fusion' where we add an additional segmentation head to both 2D and 3D network streams \textit{prior} to the fusion layer with the purpose of mimicking the central fusion head (see Fig.\,~\ref{fig:architectureFusionXmuda}). Note that this idea could also be applied on top of other fusion architectures.

\begin{table}
\centering
\resizebox{.45\textwidth}{!}{%
    \begin{tabular}{llcc}
        \toprule
        Method & Arch. & \begin{tabular}{@{}c@{}}nuSc-Lidarseg: \\  USA/Singap.\end{tabular}  & \begin{tabular}{@{}c@{}}A2D2/ \\  Sem.KITTI\end{tabular} \\
        \midrule
        Baseline (src only) & Vanilla & 66.5 & 34.2 \\
        \midrule
        Deep logCORAL~\cite{morerio2017minimal} & Vanilla & 64.0 & 36.2 \\
        MinEnt~\cite{vu2019advent} & Vanilla & 65.4 & 39.8 \\
            PL~\cite{li2019bidirectional} & Vanilla & \sndbest{70.1} & 38.6 \\
        \midrule
                xMUDA Fusion & xMUDA & 69.3 & \textbf{42.6} \\
                xMUDA\textsubscript{PL} Fusion & xMUDA & \textbf{70.7} & \sndbest{42.2} \\
        \midrule\midrule
        Oracle & xMUDA & 80.6 & 65.7 \\
        \bottomrule
    \end{tabular}
    }
	\caption{\textbf{Comparison of the fusion methods}. Performance in mIoU for the two UDA scenarios: nuScenes-Lidarseg~\cite{nuscenes2019}: USA/Singapore and A2D2\cite{aev2019}/SemanticKITTI\cite{behley2019iccv}. We adapt the supervised baseline `Baseline (src only)' and the UDA baselines (`Deep logCORAL', `MinEnt', `PL') to the vanilla fusion architecture depicted in Fig.~\ref{fig:architectureFusionVanilla}. We propose `xMUDA Fusion' which uses the architecture of Fig.~\ref{fig:architectureFusionXmuda}.}
	\label{tab:fusion}
\end{table}

In Tab.\,~\ref{tab:fusion} we show results for different fusion approaches where we specify which architecture was used (Vanilla late fusion from Fig.\,~\ref{fig:architectureFusionVanilla} or xMUDA Fusion from Fig.\,~\ref{fig:architectureFusionXmuda}).
We observe that the xMUDA fusion architecture leads to better results than the UDA baselines with the Vanilla architecture. This demonstrates how cross-modal learning can be applied effectively in fusion setups.

\section{Ablation Studies}

\subsection{Single vs. Dual Segmentation Head}
\label{sec:dual_head_abl_study}
\begin{figure}
	\centering
	\includegraphics[width=0.5\linewidth]{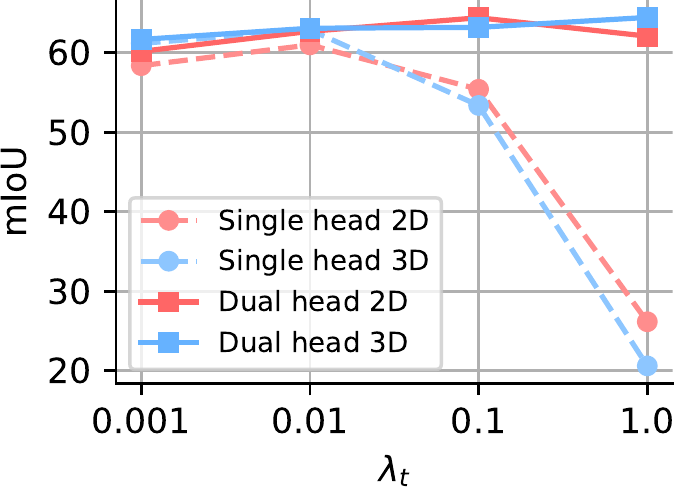}
	\caption{\textbf{Single vs. Dual Head Architecture}. mIoU of both architectures on nuScenes-Lidarseg~\cite{nuscenes2019}: USA/Singapore for different values of the target loss weight $\lambda_t$, with $\lambda_s = 1.0$.}
	\label{fig:singleVsDualHeadResults}
\end{figure}

Here we justify our dual head over the simpler single-head architecture, shown in Fig.~\ref{fig:singleVsDualHeadArch}.
In the single-head architecture (Fig.~\ref{fig:singleHead}), the cross-modal loss $\mathcal{L}_{\text{xM}}$ is directly applied between the 2D and 3D main heads. This enforces consistency by aligning the two outputs in addition to the supervised segmentation loss $\mathcal{L}_{\text{seg}}$. Thus, the heads must satisfy the two objectives --segmentation and consistency-- at the same time. To showcase the disadvantage of this architecture, we train xMUDA (as in Eq.~\ref{eq:completeObjective}) and vary the weight $\lambda_t$ for the cross-modal loss $\mathcal{L}_{\text{xM}}$ on target, which is the main driver for UDA. The results in Fig.~\ref{fig:singleVsDualHeadResults} for the \textit{single-head} architecture (Fig.~\ref{fig:singleHead}) show that increasing $\lambda_t$ from 0.001 to 0.01 slightly improves the mIoU, but that increasing $\lambda_t$ further to 0.1 and 1.0, has a hugely negative effect. In the extreme case $\lambda_t=1.0$, 2D and 3D always predict the same class, thus only satisfying the consistency, but not the segmentation objective.

The \textit{dual-head} architecture (Fig.~\ref{fig:dualHead}) addresses this problem by introducing a secondary mimicking head which purpose is to mimic the main head of the other modality during the training and can be discarded afterwards. This effectively disentangles the mimicking objective which is applied to the mimicking head from the segmentation objective which is applied to the main head. Fig.~\ref{fig:singleVsDualHeadResults} shows that increasing $\lambda_t$ to 0.1 for \textit{dual-head} produces the best results overall --better than any value for $\lambda_t$ for \textit{single head}-- and that the results are robust even at $\lambda_t=1.0$.

\subsection{Cross-Modal Learning on Source}

In Eq.~\ref{eq:completeObjective}, the cross-modal loss $\mathcal{L}_{\text{xM}}$ is applied on source \textit{and} target domains, although we already have the supervised segmentation loss $\mathcal{L}_{\text{seg}}$ on source domain. We observe a gain of 1.6 mIoU on 2D and 1.3 on 3D when adding $\mathcal{L}_{\text{xM}}$ on source domain as opposed to applying it on target domain \textit{only}. This shows that it is important to train the mimicking head on source-domain data, stabilizing predictions, which can be exploited during adaptation on target-domain inputs. 

\subsection{Cross-modal Supervised Learning }

\begin{table}
\setlength\heavyrulewidth{0.25ex}
\centering
\resizebox{.45\textwidth}{!}{
        \begin{tabular}{lcccccc}
        \toprule
        & \multicolumn{3}{c}{nuScenes-Lidarseg:\,Singap. ($\mathcal{T}$)} & \multicolumn{3}{c}{Waymo\,OD: KRK ($\mathcal{T}_{\ell}$)} \\
        \cmidrule(r){2-4}
        \cmidrule(r){5-7}
        Loss & 2D & 3D & 2D+3D & 2D & 3D & 2D+3D \\
        \midrule
        $\mathcal{L}_{\text{seg}}$ & 75.4 & 76.0 & 79.6 & 51.3 & \textbf{57.7} & 59.2 \\
        $\mathcal{L}_{\text{seg}} + \mathcal{L}_{\text{xM}}$ & \textbf{75.8} & \textbf{76.2} & \textbf{79.8} & \textbf{57.4} & 57.6 & \textbf{61.1} \\
        \bottomrule
        \end{tabular}}
	\caption{\textbf{Benefit of the proposed cross-modal loss in supervised learning}. Performance in mIoU of supervised learning with and without cross-modal loss $\mathcal{L}_{\text{xM}}$ on nuScenes-Lidarseg~\cite{nuscenes2019} (Singapore) and on Waymo\,OD~\cite{sun2020scalability} (KRK), using only the labeled target-domain dataset~$\mathcal{T}_{\ell}$. In Singapore experiment, the model trained with the cross-modal loss amounts to the oracle on this dataset in Tab.~\protect{\ref{tab:resultsXmuda}}.}
	\label{tab:xmSupervisedResults}
\end{table}

To evaluate the possible benefits of cross-modal learning for purely supervised settings, we conducted experiments with and without adding the cross-modal loss $\mathcal{L}_{\text{xM}}$ on two different target-domain datasets: nuScenes-Lidarseg~\cite{nuscenes2019} and Waymo\,OD~\cite{sun2020scalability}. The results are shown in Tab.~\ref{tab:xmSupervisedResults} and show a performance gain when adding $\mathcal{L}_{\text{xM}}$.
We hypothesize that the extra cross-modal objective can be beneficial, similar to multi-task learning. On the Waymo\,OD dataset, we observe a strong improvement on 2D. We observe in the training curve (validation) that cross-modal learning reduces overfitting in 2D. We hypothesize that 3D, which suffers less from overfitting, can have a regularizing effect on 2D.

\section{Conclusion}

In this work, we proposed cross-modal learning for domain adaptation in unsupervised (xMUDA) and semi-supervised (\xMSSDA{}) settings. To this end, we designed a two-stream, dual-head architecture and applied a cross-modal loss to the image and point-cloud modalities in the task of 3D semantic segmentation. The cross-modal loss consists of KL divergence applied between the predictions of the two modalities and thereby enforces consistency.

Experiments on four unsupervised and four semi-supervised domain adaptation scenarios show that cross-modal learning outperforms uni-modal adaptation baselines and is complementary to learning with pseudo-labels.

We think that cross-modal learning could generalize to many tasks that involve multi-modal input data and is not constrained to DA or to image and point-cloud modalities.



%



\appendices

Here we provide more details of the dataset splits used in our experiments and additional qualitative results.

\section{Dataset Splits}\label{sec:appendix:datasetSplits}
\subsection{nuScenes (UDA)}\label{sec:nuscenes}
The nuScenes dataset~\cite{nuscenes2019} consists of 1000 driving scenes, each of 20 seconds, which corresponds to 40k annotated keyframes taken at 2Hz. The scenes are split into train (28,130 keyframes), validation (6,019 keyframes) and hidden test set. The point-wise 3D semantic labels are provided by nuScenes-Lidarseg. We propose the following splits destined for domain adaptation with the respective source/target domains: Day/Night and Boston/Singapore. Therefore, we use the official validation split as test set and divide the training set into train/val for the target set. As the number of points in the target split (e.g. for night) can be very small for some classes, we group the classes bicycle, bus, car, construction vehicle, motorcycle, trailer, truck under \textbf{vehicle}, ignore the classes barrier, pedestrian, traffic cone, other flat, but keep the classes \textbf{driveable surface}, \textbf{sidewalk}, \textbf{terrain}, \textbf{manmade} and \textbf{vegetation} as is. Hence, there are 6 classes in total after the class mapping.

\subsection{VirtualKITTI/SemanticKITTI (UDA)}\label{sec:virtKittiSemKitti}

VirtualKITTI (v.1.3.1) \cite{gaidon2016virtual} consists of 5 driving scenes which were created with the Unity game engine by real-to-virtual cloning of the scenes 1, 2, 6, 18 and 20 of the real KITTI dataset~\cite{geiger2012cvpr}, i.e. bounding box annotations of the real dataset were used to place cars in the virtual world. Different from real KITTI, VirtualKITTI does not simulate LiDAR, but rather provides a dense depth map, alongside semantic, instance and flow ground truth. Each of the 5 scenes contains between 233 and 837 frames, i.e. in total 2126 for the 5 scenes. Each frame is rendered with 6 different weather/lighting variants (clone, morning, sunset, overcast, fog, rain) which we use all. Note that we do not use the renderings with different horizontal rotations. We use the whole VirtualKITTI dataset as source training set.

The SemanticKITTI dataset \cite{behley2019iccv} provides 3D point cloud labels for the Odometry dataset of KITTI~\cite{geiger2012cvpr} which features large-angle front camera and a 64-layer LiDAR. The annotation of the 28 classes has been carried out directly in 3D.

We use scenes $\{0, 1, 2, 3, 4, 5, 6, 9, 10\}$ as train set, $7$ as validation and $8$ as test set.

We select 6 shared classes between the 2 datasets by merging or ignoring them (see Tab.~\ref{tab:classMappingVirtualKittiSemKitti}). The 6 final classes are vegetation\_terrain, building, road, object, truck, car.

\begin{table*}
	\scriptsize
	\centering
	\begin{tabular}{ll|ll}
        class VirtualKITTI & mapped class & class SemanticKITTI & mapped class \\
        \toprule
        Terrain & vegetation\_terrain & unlabeled & ignore \\
        Tree & vegetation\_terrain & outlier & ignore \\
        Vegetation & vegetation\_terrain & car & car \\
        Building & building & bicycle & ignore \\
        Road & road & bus & ignore \\
        TrafficSign & object & motorcycle & ignore \\
        TrafficLight & object & on-rails & ignore \\
        Pole & object & truck & truck \\
        Misc & object & other-vehicle & ignore \\
        Truck & truck & person & ignore \\
        Car & car & bicyclist & ignore \\
        Van & ignore & motorcyclist & ignore \\
        Don't care & ignore & road & road \\
        & & parking & ignore \\
        & & sidewalk & ignore \\
        & & other-ground & ignore \\
        & & building & building \\
        & & fence & object \\
        & & other-structure & ignore \\
        & & lane-marking & road \\
        & & vegetation & vegetation\_terrain \\
        & & trunk & vegetation\_terrain \\
        & & terrain & vegetation\_terrain \\
        & & pole & object \\
        & & traffic-sign & object \\
        & & other-object & object \\
        & & moving-car & car \\
        & & moving-bicyclist & ignore \\
        & & moving-person & ignore \\
        & & moving-motorcyclist & ignore \\
        & & moving-on-rails & ignore \\
        & & moving-bus & ignore \\
        & & moving-truck & truck \\
        & & moving-other-vehicle & ignore \\
        \bottomrule
\end{tabular}
	\caption{Class mapping for VirtualKITTI/SemanticKITTI UDA scenario.}
	\label{tab:classMappingVirtualKittiSemKitti}
\end{table*}

\subsection{A2D2/SemanticKITTI (UDA+SSDA)}

The A2D2 dataset \cite{aev2019} features 20 drives, which corresponds to 28,637 frames. The point cloud comes from three 16-layer front LiDARs (left, center, right) where the left and right front LiDARS are inclined. The semantic labeling was carried out in the 2D image for 38 classes and we compute the 3D labels by projection of the point cloud into the labeled image. We keep scene 20180807\_145028 as test set and use the rest for training.

Please refer to Sec.~\ref{sec:virtKittiSemKitti} for details on SemanticKITTI. For UDA, we use the same split as in VirtualKITTI/SemanticKITTI, i.e. scenes $\{0, 1, 2, 3, 4, 5, 6, 9, 10\}$ as train set, $7$ as validation and $8$ as test set. For SSDA, we use the scenes $\{0, 1\}$ as labeled train set $\mathcal{T}_{\ell}$, $\{2\dotsc6, 9\dotsc21\}$ as unlabeled train set $\mathcal{T}_u$, $7$ as validation and $8$ as test set.

We select 10 shared classes between the 2 datasets by merging or ignoring them (see Tab.~\ref{tab:classMappingA2D2SemKitti}). The 10 final classes are car, truck, bike, person, road, parking, sidewalk, building, nature, other-objects.

\begin{table*}
	\scriptsize
	\centering
	\begin{tabular}{ll|ll}
    A2D2 class & mapped class & SemanticKITTI class & mapped class \\
    \toprule
    Car 1 & car & unlabeled & ignore \\
    Car 2 & car & outlier & ignore \\
    Car 3 & car & car & car \\
    Car 4 & car & bicycle & bike \\
    Bicycle 1 & bike & bus & ignore \\
    Bicycle 2 & bike & motorcycle & bike \\
    Bicycle 3 & bike & on-rails & ignore \\
    Bicycle 4 & bike & truck & truck \\
    Pedestrian 1 & person & other-vehicle & ignore \\
    Pedestrian 2 & person & person & person \\
    Pedestrian 3 & person & bicyclist & bike \\
    Truck 1 & truck & motorcyclist & bike \\
    Truck 2 & truck & road & road \\
    Truck 3 & truck & parking & parking \\
    Small vehicles 1 & bike & sidewalk & sidewalk \\
    Small vehicles 2 & bike & other-ground & ignore \\
    Small vehicles 3 & bike & building & building \\
    Traffic signal 1 & other-objects & fence & other-objects \\
    Traffic signal 2 & other-objects & other-structure & ignore \\
    Traffic signal 3 & other-objects & lane-marking & road \\
    Traffic sign 1 & other-objects & vegetation & nature \\
    Traffic sign 2 & other-objects & trunk & nature \\
    Traffic sign 3 & other-objects & terrain & nature \\
    Utility vehicle 1 & ignore & pole & other-objects \\
    Utility vehicle 2 & ignore & traffic-sign & other-objects \\
    Sidebars & other-objects & other-object & other-objects \\
    Speed bumper & other-objects & moving-car & car \\
    Curbstone & sidewalk & moving-bicyclist & bike \\
    Solid line & road & moving-person & person \\
    Irrelevant signs & other-objects & moving-motorcyclist & bike \\
    Road blocks & other-objects & moving-on-rails & ignore \\
    Tractor & ignore & moving-bus & ignore \\
    Non-drivable street & ignore & moving-truck & truck \\
    Zebra crossing & road & moving-other-vehicle & ignore \\
    Obstacles / trash & other-objects & & \\
    Poles & other-objects & & \\
    RD restricted area & road & & \\
    Animals & other-objects & & \\
    Grid structure & other-objects & & \\
    Signal corpus & other-objects & & \\
    Drivable cobbleston & road & & \\
    Electronic traffic & other-objects & & \\
    Slow drive area & road & & \\
    Nature object & nature & & \\
    Parking area & parking & & \\
    Sidewalk & sidewalk & & \\
    Ego car & car & & \\
    Painted driv. instr. & road & & \\
    Traffic guide obj. & other-objects & & \\
    Dashed line & road & & \\
    RD normal street & road & & \\
    Sky & ignore & & \\
    Buildings & building & & \\
    Blurred area & ignore & & \\
    Rain dirt & ignore & & \\
    \bottomrule
\end{tabular}
	\caption{\textbf{Class mapping} for A2D2/SemanticKITTI UDA and SSDA scenario.}
	\label{tab:classMappingA2D2SemKitti}
\end{table*}

\subsection{Waymo OD (SSDA)}

The Waymo Open Dataset (v.1.2.0) provides 1150 scenes of 20s each. For simplicity and consistency with other UDA scenarios, we only use the top, but not the 4 side LiDARs, and only the front, but not the 4 side cameras. Similar to nuScenes (Sec.~\ref{sec:nuscenes}), we obtain segmentation labels from 3D bounding boxes.

There is a main dataset which we use as source dataset and a partially labeled domain adaptation dataset of which we use the labeled part as labeled target set $\mathcal{T}_{\ell}$ and the unlabeled part as unlabeled target set $\mathcal{T}_u$.

We ignore the cyclist class, because there are no cyclist labels available in the target data, i.e. we only keep the classes vehicle, pedestrian, sign, unknown.

\section{Opposite adaptation direction}\label{sec:appendix:oppositeDir}
In order to test if our proposed method also works in the opposite direction, i.e. Night/Day instead of Day/Night, we run the experiments for `Baseline (src only)' and our methods `xMUDA' and `xMUDA\textsubscript{PL}' on nuScenes-Lidarseg: Night/Day and Singapore/USA, and report the results in Tab.~\ref{tab:oppositeDirection}.
The results show no surprises. When comparing 2D+3D results, our xMUDA\textsubscript{PL} improves mIoU by +3.5 and +8.8 mIoU with respect to `Baseline (src only)', on Singapore/USA and Night/Day, respectively. These new results are aligned -- if not better -- with their xMUDA\textsubscript{PL} counterparts from Tab.~2 of the main paper (+3 and +1.1 mIoU on USA/Singapore and Day/Night, respectively).

\begin{table}
\centering
    \resizebox{.47\textwidth}{!}{%
    \begin{tabular}{lcccccc}
    \toprule
    & \multicolumn{3}{c}{nuSc-Lidarseg: Singap./USA} & \multicolumn{3}{c}{nuSc-Lidarseg: Night/Day} \\
    \cmidrule(r){2-4}
    \cmidrule(r){5-7}
    Method & 2D & 3D & 2D+3D & 2D & 3D & 2D+3D \\
    \midrule
    Baseline (src only) & 62.2 & 68.4 & 71.3 & 55.1 & 70.3 & 64.7 \\
    \midrule
    xMUDA & 69.2 & 70.0 & 73.5 & 67.4 & 71.1 & 71.9 \\
    xMUDA\textsubscript{PL} & \textbf{70.8} & \textbf{73.0} & \textbf{74.8} & \textbf{68.9} & \textbf{72.6} & \textbf{73.5} \\
    \bottomrule
    \end{tabular}}
	\caption{\textbf{Opposite UDA performance of xMUDA on nuScenes-Lidarseg~\cite{nuscenes2019}.} We report the `Baseline (src only)', xMUDA and xMUDA\textsubscript{PL}, observing that our proposals are also effective in the opposite adaptation direction, i.e. Singapore/USA instead of USA/Singapore in the main experiments and analogous for Night/Day instead of Day/Night.
	}
	\label{tab:oppositeDirection}
\end{table}

\section{3D UDA baseline: LiDAR transfer}\label{sec:appendix:lidarTransfer}

LiDAR transfer~\cite{langer2020domain} is a 3D UDA baseline that aligns source and target LiDAR data in point-cloud input space. It is specifically designed to adapt from 64-layer LiDAR data to 32-layer LiDAR, i.e. from high to low LiDAR resolution.

In the following we evaluate LiDAR transfer~\cite{langer2020domain} and xMUDA/xMUDA\textsubscript{PL} on the UDA scenario SemanticKITTI/nuScenes proposed in~\cite{langer2020domain}. To accommodate to our setup, we replaced the 2D CNN from~\cite{langer2020domain} by a sparse 3D CNN~\cite{SparseConvNet} and applied a class mapping resulting in 6 classes (vehicle, driveable surface, sidewalk, terrain, manmade and vegetation), as can be seen in Tab.~\ref{tab:classMappingSemKITTInuScenesLidarseg}.

In Tab.~\ref{tab:lidarTransfer}, we provide the results of `Baseline (src only)', `LiDAR transfer (re-style)' which is trained on the downsampled (64 to 32 layers) source dataset, `LiDAR transfer (full)' which additionally applies Deep logCORAL~\cite{morerio2017minimal} and finally our methods xMUDA and xMUDA\textsubscript{PL}.

We observe that both LiDAR transfer methods (re-style and full) perform worse than `Baseline (src only)', while xMUDA and xMUDA\textsubscript{PL} have a considerable domain adaptation effect.

\begin{table}
\centering
    \begin{tabular}{lccc}
    \toprule
    & \multicolumn{3}{c}{Sem.KITTI/nuSc-Lidarseg} \\
    \cmidrule(r){2-4}
    Method & 2D & 3D & 2D+3D \\
    \midrule
    Baseline (src only) & 47.6 & 54.9 & 61.5 \\
    \midrule
    LiDAR transfer (re-style)~\cite{langer2020domain} & - & 53.1 & - \\
    LiDAR transfer (full)~\cite{langer2020domain} & - & 54.2 & - \\
    \midrule
    xMUDA & 57.6 & 57.7 & 63.2 \\
    xMUDA\textsubscript{PL} & 61.0 & 61.4 & 67.5 \\
    \bottomrule
    \end{tabular}
	\caption{\textbf{Comparison of Lidar Transfer~\cite{langer2020domain} with xMUDA}. Performance in mIoU for the UDA scenario: SemanticKITTI/nuScenes-Lidarseg. `LiDAR transfer (re-style)~\cite{langer2020domain}' restyles the source (64 layers) as target by sub-sampling (32 layers) and `LiDAR transfer (full)~\cite{langer2020domain}' adds Deep logCORAL~\cite{morerio2017minimal} on top.
	}
	\label{tab:lidarTransfer}
\end{table}

\begin{table*}
\centering
    \begin{tabular}{ll|ll}
    SemanticKITTI class & mapped class & nuScenes-LidarSeg class & mapped class \\
    \toprule
    unlabeled & ignore & ignore & ignore \\
    outlier & ignore & barrier & ignore \\
    car & vehicle & bicycle & vehicle \\
    bicycle & vehicle & bus & vehicle \\
    bus & ignore & car & vehicle \\
    motorcycle & vehicle & construction\_vehicle & vehicle \\
    on-rails & ignore & motorcycle & vehicle \\
    truck & vehicle & pedestrian & ignore \\
    other-vehicle & ignore & traffic\_cone & ignore \\
    person & ignore & trailer & vehicle \\
    bicyclist & vehicle & truck & vehicle \\
    motorcyclist & vehicle & driveable\_surface & driveable\_surface \\
    road & driveable\_surface & other\_flat & ignore \\
    parking & driveable\_surface & sidewalk & sidewalk \\
    sidewalk & sidewalk & terrain & terrain \\
    other-ground & ignore & manmade & manmade \\
    building & manmade & vegetation & vegetation \\
    fence & manmade & & \\
    other-structure & ignore & & \\
    lane-marking & driveable\_surface & & \\
    vegetation & vegetation & & \\
    trunk & vegetation & & \\
    terrain & terrain & & \\
    pole & manmade & & \\
    traffic-sign & manmade & & \\
    other-object & manmade & & \\
    moving-car & vehicle & & \\
    moving-bicyclist & vehicle & & \\
    moving-person & ignore & & \\
    moving-motorcyclist & vehicle & & \\
    moving-on-rails & ignore & & \\
    moving-bus & ignore & & \\
    moving-truck & vehicle & & \\
    moving-other-vehicle & ignore & & \\
    \bottomrule
    \end{tabular}
	\caption{\textbf{Class mapping for SemanticKITTI/nuScenes-Lidarseg UDA scenario}. Note that the mapping on nuScenes-Lidarseg (right side of the table) is the same as described in Sec.~\ref{sec:nuscenes}.
	}
	\label{tab:classMappingSemKITTInuScenesLidarseg}
\end{table*}

\section{Additional qualitative Results (UDA)}\label{sec:appendix:additionalQualRes}
\begin{figure*}
	\centering
	\includegraphics[width=0.5\textwidth]{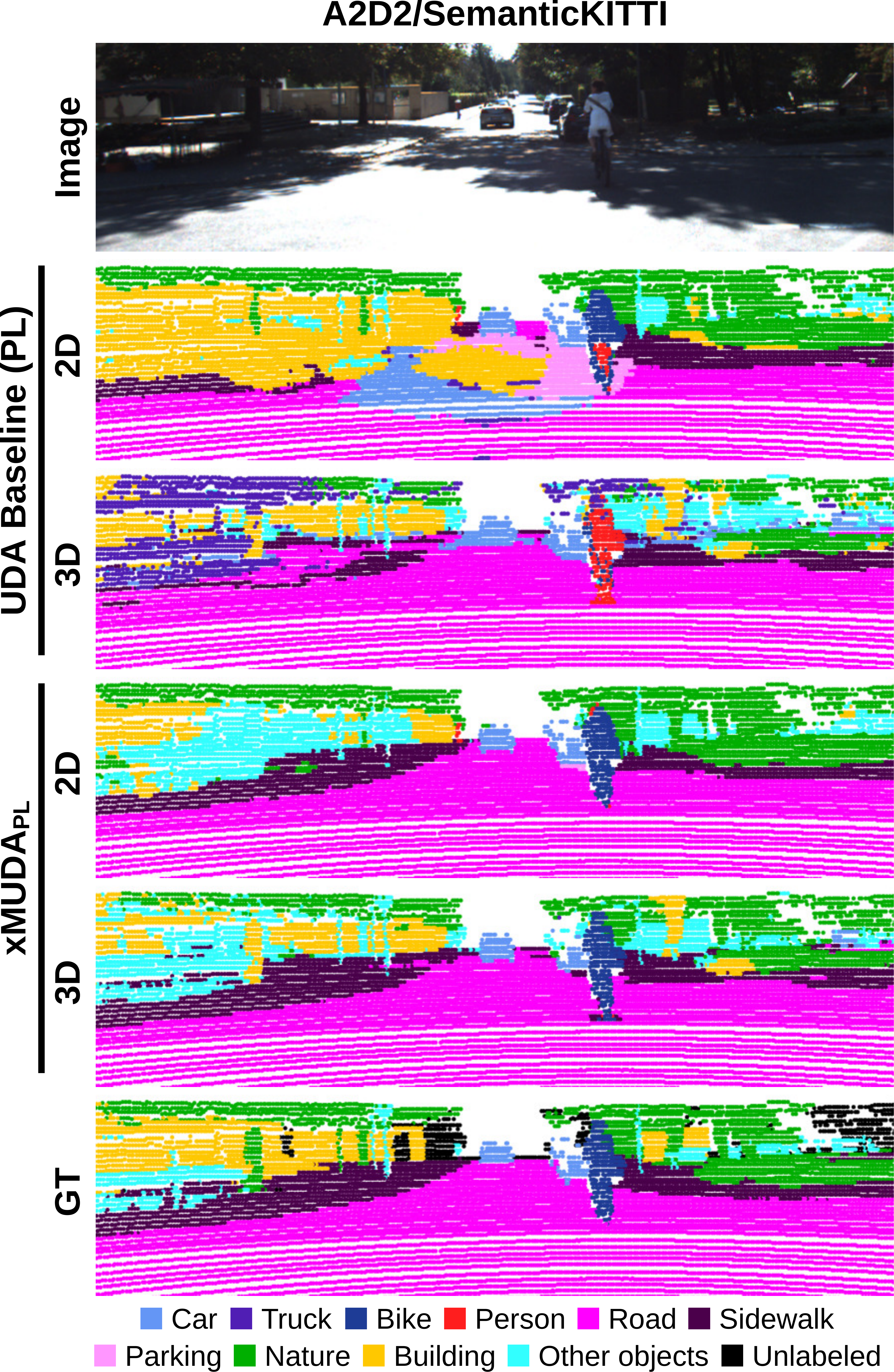}
	\caption{\textbf{Qualitative results on A2D2/SemanticKITTI (UDA)}. For UDA Baseline (PL) and xMUDA\textsubscript{PL}, we separately show the predictions of the 2D and 3D network stream.\\
	For the uni-modal UDA baseline (PL), the 2D prediction lacks consistency on the road and 3D is unable to recognize the bike and the building on the left correctly. In xMUDA\textsubscript{PL}, both modalities can stabilize each other and obtain better performance on the bike, the road, the sidewalk and the building.
	}
	\label{fig:qualitativeResultsSupp}
\end{figure*}

We provide additional qualitative results for UDA for the scenarios nuScenes:\,Day/Night and A2D2/SemanticKITTI in Fig.~\ref{fig:qualitativeResultsSupp}, where we show the output of the 2D and 3D stream individually to illustrate their respective strengths and weaknesses, e.g. that 3D works much better at night.

\section{SSDA example: no gain from source data}\label{sec:appendix:noGainFromSour}

\begin{table}
\centering
\resizebox{\columnwidth}{!}{
	\begin{threeparttable}
        \begin{tabular}{llccc}
        \toprule
        & & \multicolumn{3}{c}{Virt.KITTI/Sem.KITTI} \\
        \cmidrule(r){3-5}
        Method & Train set & 2D & 3D & 2D+3D \\
        \midrule
        Baseline (src only) & $\mathcal{S}$ & 26.8 & 42.0 & 42.2 \\
        Baseline (lab. trg only) & $\mathcal{T}_\ell$ & \textbf{63.4} & 69.3 & 71.2 \\
        Baseline (src and lab. trg) & $\mathcal{S}$~+~$\mathcal{T}_\ell$ & 62.7 & 71.5 & 71.0 \\
        \midrule\midrule
        Domain gap ($\mathcal{S}$ vs. $\mathcal{S}$ + $\mathcal{T}_{\ell}$) & & 35.9 & 29.5 & 28.8 \\
        \midrule
        xMUDA & $\mathcal{S}$~+~$\mathcal{T}_u$ & 46.7 & 46.1 & 52.4 \\
        xMUDA\textsubscript{PL} & $\mathcal{S}$~+~$\mathcal{T}_u$ & 49.5 & 54.6 & 54.4 \\
        \midrule
        Deep logCORAL~\cite{morerio2017minimal} & $\mathcal{S}$~+~$\mathcal{T}_\ell$~+~$\mathcal{T}_u$ & 61.7\tnote{*} & 69.4 & 70.1 \\
        MinEnt~\cite{vu2019advent} & $\mathcal{S}$~+~$\mathcal{T}_\ell$~+~$\mathcal{T}_u$ & 61.1 & 71.0 & 70.6 \\
        PL~\cite{li2019bidirectional} & $\mathcal{S}$~+~$\mathcal{T}_\ell$~+~$\mathcal{T}_u$ & 62.1 & 71.4 & 70.9 \\
        \midrule
        \xMSSDA{} & $\mathcal{S}$~+~$\mathcal{T}_\ell$~+~$\mathcal{T}_u$ & \sndbest{63.0} & \sndbest{72.9} & \textbf{72.2} \\
        \xMSSDA{}\textsubscript{PL} & $\mathcal{S}$~+~$\mathcal{T}_\ell$~+~$\mathcal{T}_u$ & 62.8 & \textbf{76.1} & \sndbest{71.4} \\
        \midrule\midrule
        Unsupervised advantage & & 0.1 & 4.6 & 0.4 \\
        \tiny{(relative)} & & \tiny{(+0.2\%)} & \tiny{(+6.4\%)} & \tiny{(+0.5\%)} \\
        \bottomrule
        \end{tabular}
	\begin{tablenotes}
	\item[*] The 2D network is trained with batch size 6 instead of 8 to fit into GPU memory.
	\end{tablenotes}
	\end{threeparttable}}
	\caption{\textbf{\xMSSDA{} experiments for the SSDA scenario VirtualKITTI~\cite{gaidon2016virtual}/SemanticKITTI~\cite{behley2019iccv} where the benefit of the source dataset $\mathcal{S}$ is arguable}. As in the main paper in Tab.~3, we report the mIoU result (with \best{best} and \sndbest{2nd best}) on the target set for each network stream (2D and 3D) as well as the ensembling result taking the mean of the 2D and 3D probabilities (2D+3D).}
	\label{tab:extraXmssda}
\end{table}

Before applying domain adaptation techniques, it is beneficial to first study the data situation.

In the UDA case, when no labeled target data is available, even a small amount of labeled source data represents a great benefit, because labels on a different domain are still better than no labels at all.

However, the SSDA case is more complex. As a small amount of labeled target data $\mathcal{T}_\ell$ is available, one must rely less on the (labeled) source dataset $\mathcal{S}$. The impact of domain adaptation is most significant when the source-target domain gap is small and the amount of labeled source data $\mathcal{S}$ is significant in comparison to the amount of labeled target data $\mathcal{T}_\ell$.

It is impossible to say in general when the inclusion of source data is beneficial. However, in this section, we still want to give an example to the reader where the inclusion of source data for the training is arguable. We present the results of the VirtualKITTI~\cite{gaidon2016virtual}/SemanticKITTI~\cite{behley2019iccv} SSDA scenario in Tab.~\ref{tab:extraXmssda}.

In this scenario, the number of frames in the training sets accounts to 2,126 in the source dataset $\mathcal{S}$, to 5,642 in the labeled target dataset $\mathcal{T}_\ell$ and to 32,738 in the unlabeled target dataset $\mathcal{T}_u$. Hence, there are 2.65x more labeled frames in $\mathcal{T}_\ell$ than in $\mathcal{S}$ which poses the question if the small source dataset $\mathcal{S}$ is useful.

Tab.~\ref{tab:extraXmssda} shows that for 2D, `Baseline (lab. trg only)` trained only on $\mathcal{T}_\ell$ performs better (63.4 mIoU) than `Baseline (src and lab. trg)` trained on $\mathcal{S}$~+~$\mathcal{T}_\ell$ (62.7 mIoU). Hence, the inclusion of source training data harms performance. We hypothesize that the mIoU decrease stems from the large (2D) image virtual-to-real domain gap. Instead, for 3D, we observe a small improvement, i.e. from 69.3 mIoU, training on $\mathcal{T}_\ell$, to 71.5 mIoU, training on $\mathcal{S}$~+~$\mathcal{T}_\ell$. This might be thanks to the fact that the 3D point cloud domain gap (29.5) is smaller than for 2D images (35.9).

Still, \xMSSDA{} outperforms baselines~\cite{morerio2017minimal,vu2019advent,li2019bidirectional} showing the benefit of cross-modal learning on $\mathcal{T}_u$.



\ifCLASSOPTIONcaptionsoff
  \newpage
\fi



%



\bibliographystyle{IEEEtran}
\bibliography{refBase}

%
\newcommand{\vspacelength}{-47pt}

\vspace{-2em}
\begin{IEEEbiography}[{\includegraphics[width=1in,height=1.10in,clip,keepaspectratio]{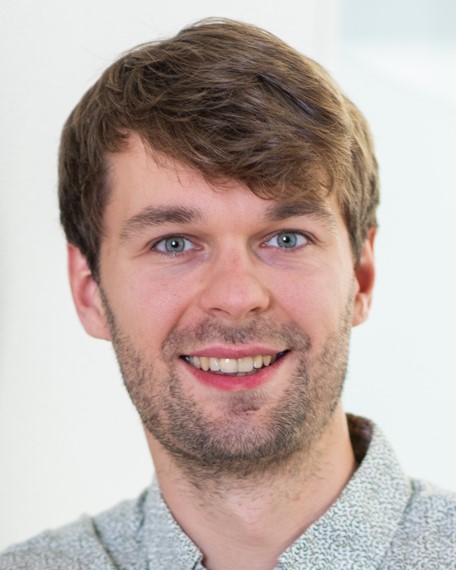}}]{Maximilian Jaritz} is an Applied Scientist at Amazon Robotics AI since 2020. Before, he was a PhD student with Inria and Valeo/Valeo.ai under the supervision of Raoul de Charette, Emilie Wirbel and Patrick P\'rez. He has obtained Master's degrees in a dual degree program from TU Berlin and Ecole Centrale Paris in 2016.
\end{IEEEbiography}

\vspace{\vspacelength}
\begin{IEEEbiography}[{\includegraphics[width=1in,height=1.10in,clip,keepaspectratio]{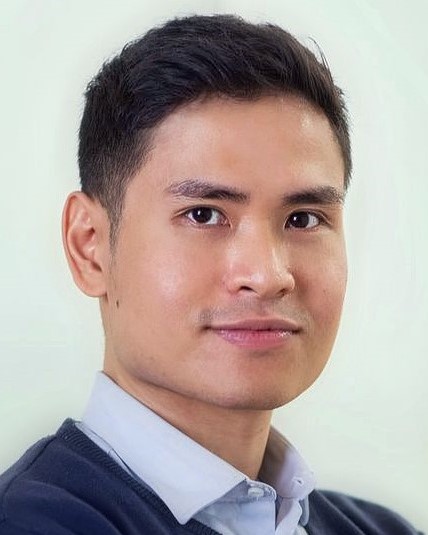}}]{Tuan-Hung Vu} is a research scientist at Valeo.ai since 2018. He received an engineering master degree in machine learning from T\'el\'ecom Paristech and a PhD degree in Computer Vision from \'Ecole Normale Sup\'erieure Paris. His research interests include deep learning, scene understanding, domain adaptation and data augmentation.
\end{IEEEbiography}

\vspace{\vspacelength}
\begin{IEEEbiography}[{\includegraphics[width=1in,height=1.10in,clip,keepaspectratio]{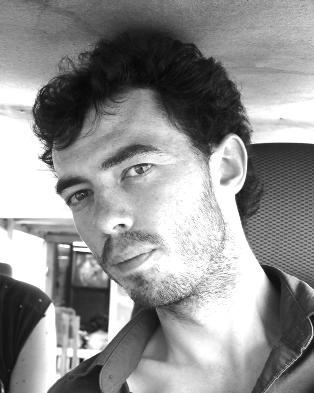}}]{Raoul de Charette} is a researcher at Inria Paris since 2015. He received the MSc degree in arts and computer graphics from Paris 8 Uni. and the PhD degree in computer vision from Mines ParisTech in 2012. He was consecutively with Mines ParisTech (2008-2011, 2012-2013), Carnegie Mellon University (2011), and University of Makedonia (2014). He now leads the Computer Vision Group in RITS Team Inria. 
\end{IEEEbiography}

\vspace{\vspacelength}
\begin{IEEEbiography}[{\includegraphics[width=1in,height=1.10in,clip,keepaspectratio]{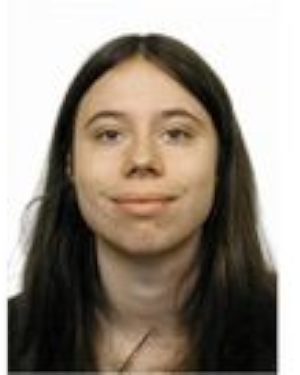}}]{\'Emilie Wirbel} is a Senior Robotics Engineer at NVIDIA. Before, she was with Valeo Driving Assistance Research and Valeo.ai, working on prediction and control for autonomous driving, in particular relying on end-to-end imitation learning. She received an engineering degree from Mines ParisTech in 2011, then a PhD in robotics in 2014. 
\end{IEEEbiography}

\vspace{\vspacelength}
\begin{IEEEbiography}[{\includegraphics[width=1in,height=1.10in,clip,keepaspectratio]{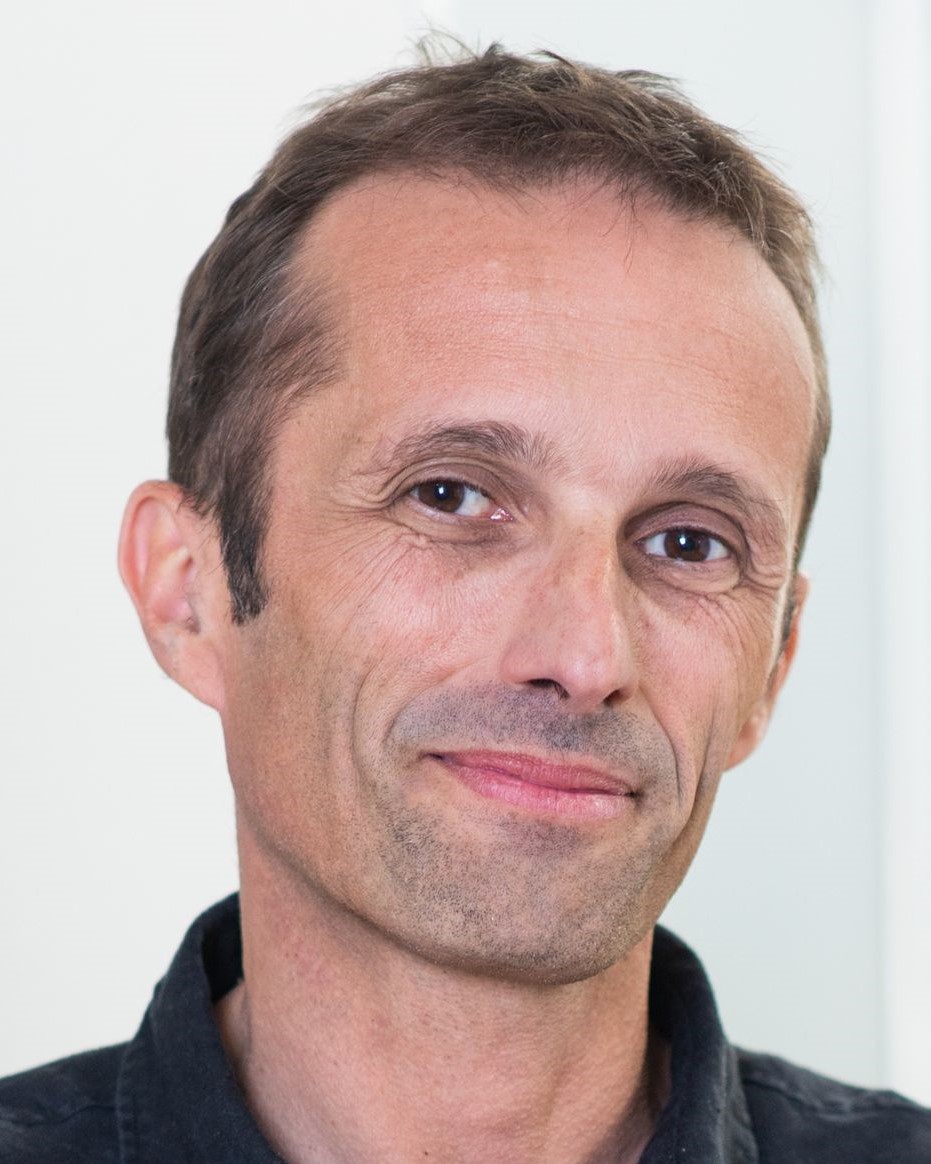}}]{Patrick P\'erez} is Scientific Director of Valeo.ai, a Valeo research lab on artificial intelligence for automotive applications. Before joining Valeo, Patrick P\'erez has been Distinguished Scientist at Technicolor (2009-2918), researcher at Inria (1993-2000, 2004-2009) and at Microsoft Research Cambridge (2000-2004). 
His research revolves around machine learning for scene understanding, data mining and visual editing.
\end{IEEEbiography}




\end{document}